\documentclass[preprint]{elsarticle}
\usepackage{ifpdf}
\usepackage{cite}
\usepackage{graphicx}
\usepackage{epstopdf}
\usepackage{amssymb}
\usepackage{bm}
\usepackage{amsthm}
\usepackage[ruled,vlined]{algorithm2e}
\usepackage[caption=false, font=footnotesize]{subfig}
\usepackage{color}

\newtheorem{mydef}{Definition}

\usepackage[cmex10]{amsmath}

\begin{document}

\begin{frontmatter}

\title{Automated Peer-to-peer Negotiation for Energy Contract Settlements in Residential Cooperatives\footnote{This article extends an abridged version, which has been published and presented at the 2018 IEEE International Conference on Communications, Control, and Computing Technologies for Smart Grids~\citep{chakra2018-1}.}}

\author[UNIMELB]{Shantanu Chakraborty\corref{cor1}}
\ead{shantanu.chakraborty@unimelb.edu.au}

\author[CWI,UU]{Tim Baarslag}
\author[CWI]{Michael Kaisers}

\cortext[cor1]{Corresponding author}

\address[UNIMELB]{Energy Transition Hub, The University of Melbourne, Australia.}
\address[CWI]{Intelligent and Autonomous Systems Group, Centrum Wiskunde \& Informatica, Amsterdam, The Netherlands.}
\address[UU]{Intelligent Systems, Utrecht University, The Netherlands.}



\begin{abstract}
This paper presents an automated peer-to-peer negotiation strategy for settling \emph{energy contracts} among prosumers in a Residential Energy Cooperative considering heterogeneity prosumer preferences.
The heterogeneity arises from prosumers' evaluation of \emph{energy contracts} through multiple societal and environmental criteria and the prosumers' private preferences over those criteria.
The prosumers engage in \emph{bilateral negotiations} with peers to mutually agree on periodical \emph{energy contracts/loans} consisting of the energy volume to be exchanged at that period and the return time of the exchanged energy. 
The negotiating prosumers navigate through a common \emph{negotiation domain} consisting of potential \emph{energy contracts} and evaluate those contracts from their valuations on the entailed criteria against a \emph{utility function} that is robust against generation and demand uncertainty.
From the repeated interactions, a prosumer gradually learns about the compatibility of its peers in reaching \emph{energy contracts} that are closer to \emph{Nash} solutions. 
Empirical evaluation on real demand, generation and storage profiles -- in multiple system scales -- illustrates that the proposed negotiation based strategy can increase the system efficiency (measured by \emph{utilitarian} social welfare) and fairness (measured by \emph{Nash} social welfare) over a baseline strategy and an individual flexibility control strategy representing the status quo strategy.
We thus elicit system benefits from peer-to-peer flexibility exchange already without any central coordination and market operator, providing a simple yet flexible and effective paradigm that complements existing markets.



\end{abstract}


\end{frontmatter}

\section{Introduction}
\label{intro}
The deregulated electricity markets refrain small-scale (residential) prosumers to actively participate in the wholesale market.
The prosumers are instead serviced in retail markets, where they are individually metered by large suppliers~\citep{chao2001design}, through representative \emph{residential aggregators} and \emph{retailers}.
This situation leads to inefficiencies as the prosumers are incentivised to individually control their local energy usage without taking the overall demand and supply status into consideration.
The jointly coordinated prosumers' (distributed) energy resources potentially shape up the overall demand and offer significant value to the energy system by alleviating the need for investment in additional generation and transmission infrastructure~\citep{pudjianto2010value} and by minimizing the fluctuations due to renewable power integration~\citep{chakra2018}. 
However, properly incentivizing the prosumers to coordinate their locally owned distributed resources is quite a challenge, and justifiably a field of active research in Smart Grids. 
One avenue is to apply distributed optimization techniques that facilitate the coordination of the DERs that are owned and controlled by a single entity. 
However, these techniques are not readily applicable when these resources have different owners; at least not without considering the strategic interactions between the owners. 

Local energy exchange may offer incentives to the prosumers to engage in competition and local trading~\citep{liu2017energysharing}. For energy communities, these mechanisms may need to take into account and balance several objectives, including, next to efficiency, altruism, or fairness of allocations~\citep{cornelusse2018optimal}. 
Prosumers in an REC may have diverse preferences over how their energy profiles are valued due to various societal and environmental factors. 
For instance, the prosumers may evaluate \emph{energy contracts} based on several criteria, e.g. \emph{self-sufficiency} or \emph{autarky}, \emph{cost of energy}, \emph{loss in flexibility}, \emph{sustainability}, and so on~\citep{baarslag2017}, resulting in a private valuation.
Market-based mechanisms, in their current form, are not fully designed to handle such heterogeneity in distributed decision making, where the prosumer-specific individual allocations are of absolute necessity. In addition, for energy communities they leave open the question who executes the central market and receives central control of allocation and information. In the context of a local \emph{prosumer marketplace} with low-liquidity settings, the state of the research fails to provide any concrete guidelines on how to perform distributed energy allocations autonomously by considering prosumers' preferences. In particular, as it will be discussed in details in Section~\ref{sec_rel_works}, there is a clear research gap in addressing 
energy contract settlement for sustainability conscious prosumers who have preferences that trade-off costs with other quantifiable impacts of energy trading (e.g. \emph{autarky}, \emph{loss in flexibility}, and so on).

While complete preferences would need to be computed and revealed for market-based solutions, peer-to-peer (P2P) negotiation proceed iteratively, reducing the amount of information revealed, and keeping responsibility with the individual prosumers. Automated negotiation is an organic process of joint decision making where multiple stakeholders -- typically represented by autonomous agents -- with conflicting interests engage and make a decision~\citep{baarslag2014bid}. 
The participating agents may have the desire to cooperate in an automated negotiation setting, but due to the conflicting interests, these agents intend to join hands in reaching a common goal~\citep{jennings01a}.
The negotiation approach contrasts market-based approaches, and its iterative nature provides a more natural model for low liquidity settings, in which personalized solutions need to be found in large outcome spaces. We present an \emph{automated negotiation approach} as an energy exchange mechanism to settle energy contract as energy loans between prosumers in a P2P fashion. 
During each negotiation session, a pair of prosumers (represented by software agents) engage in a \emph{bilateral negotiation} by exchanging and eventually agreeing on \emph{energy contracts}, comprised of several \emph{negotiation issues} (here \emph{energy volume} and \emph{return time}). At the start, an agent randomly chooses its peer to engage in a negotiation, and gradually learns to make an informed peer choice by applying a learning mechanism.
The proxy agents evaluate \emph{offers} based on criteria that model the heterogeneous preferences of the users they represent: 1) \emph{loss in flexibility} (in local storage), and 2) \emph{autarky} or sustainability of the \emph{offers}. The agents can weigh these criteria differently, thereby enabling heterogeneity and trade-offs between the agents.
In addition to that, agents take into account the uncertainty imposed by demand and renewable generation prediction while iteratively offering the \emph{energy contracts} with higher \emph{utility}. While performing the repetitive interactions, the agents iteratively learn about the compatibility of other agents and consequently make an informed decision regarding choosing negotiating partners.

The main contributions of this paper are as follows.
\begin{itemize}
  \item We propose a novel and intuitive \emph{negotiation} based strategy that considers \textit{heterogeneity} in prosumers through a \emph{distributed} and \emph{autonomous} agent model, where the model encapsulates the agent's preferences on predefined criteria and uncertainties in agent's net demand in decision making process.
   \item The strategy enables the agents to gradually learn about negotiating peers and eventually to make informed peer selection that increases the overall success rate in contract settlements.
  \item We evaluate the performance of the proposed \emph{negotiation} based strategy over real residential demand, generation, and storage data to elucidate the efficiency of the strategy in increasing the social welfare over a baseline strategy and an individual flexibility control strategy.
\end{itemize}

The rest of the paper is organized as follows: Section~\ref{sec_rel_works} provides an overview of the related works and gauges the fit of the proposed strategy in a P2P local marketplace setting. Section~\ref{system_model} describes a residential prosumer model and defines the \emph{energy contract} that is used in the negotiation process. Section~\ref{sec_negotiation} presents the negotiation based energy exchange strategy, defines required algorithms and learning process, and several contextual notions of allocative efficiency. Simulation case studies are presented and discussed in Section~\ref{sec_sim}. Finally, Section~\ref{sec_conclusion} concludes the paper with a glimpse of possible follow-up research.

\section{Related Works}
\label{sec_rel_works}
In recent times, extensive research -- either as academic practices or as commercial or pilot projects -- are conducted in the area of P2P energy trading across the different value chain of the electricity grid. 
One stream of research goes into the direction of centralized or distributed coordination mechanisms of distributed energy resources (DER) to service-specific (e.g. ancillary services; voltage and frequency regulations) support or to maximize the overall economic benefits (e.g.~\citep{calvillo2016}). 
The centralized coordination mechanisms are typically performed through the direct control method via the energy storage system, electric vehicles, and thermostatically controlled loads. 
The coordinating entity, however, needs to periodically gather the states of DERs (for instance, electric vehicles~\citep{WANG20171673} and energy storage~\citep{CHAKRABORTY2016405}) to provide the optimized control signals, which becomes increasingly intractable with the scale of the distribution system (e.g. the number of energy storages in the distribution network). 
The distributed coordination mechanisms, on the other hand, iteratively seek to converge (e.g. Nash equilibrium) for a particularly desirable outcome. 
Based on the modelling paradigm of the system, several techniques, such as Lagrangian Relaxation~\citep{papadaskalopoulos2013}, and alternating direction method of multipliers~\citep{rivera2017} have been deployed for the distributed coordination. 
Both centralized and distributed coordination paradigms are interested in maximizing the overall system performance, rather than focusing on the welfare of individual prosumer. Therefore, the prosumers are often provided with additional incentive to participate in the coordination scheme.

A more centralized approach to local energy market design is to use auction formations. An auction requires to buy and sell orders of local energy submitted to a public order book. The orders are then matched either continuously~\citep{vytelingum2010} or at discrete closing times. P2P energy sharing could also facilitate the formation of a community energy market; possibly, through a community microgrid, without any centralized control~\citep{shamsi2016}). 
Most of the frameworks presented in the existing literature for P2P energy exchange are not fully adhering the P2P system architecture; they rather follow a peer-to-pool-to-peer paradigm (e.g. Fig.2. in~\citep{zhang2018}) where a centralized entity -- such as DSO -- exists to facilitate the trading between the parties, and thus are not completely automated.

Another line of research is focused on providing the P2P energy sharing platform~\citep{park2017} through the local marketplace equipped with essential functionalities where the prosumers trade or share energy with each other to achieve individual benefits~\citep{parag2016}. Unlike the aforementioned mechanisms, the prosumers are self-motivated to participate and able to exercise full control over their DERs~\citep{chao2017}.
As described in~\citep{LONG2018261}, distributed (without an intermediary) P2P energy sharing research activities are broad categories into three categories: auction model, multiagent model, and analytical model. 
In~\citep{zhou2018}, the authors develop a multiagent based simulation framework and a systemic index system for the simulation and evaluation of various P2P energy sharing mechanisms. 
A detailed four-layer system architecture model for P2P energy trading in grid-connected microgrids is proposed in~\citep{zhang2018} with the associated bidding system for the trading between prosumers and consumers. The rise of P2P interactions paradigm among different stakeholders gives rise to Game theory based strategies and frameworks that lay the foundation of distributed decision making. For instance, a coalitional game theory based approach is proposed to encourage sustainable prosumer participation in P2P energy trading is proposed in~\citep{Tushar2018P2P}. In~\citep{anoh2019}, the author optimise the social benefit of prosumers in a Virtual microgrid setting when the prosumers are exposed to different roles - producers and consumers, and analyse the allocation through Stackleberg game. The importance of applying P2P energy trading further highlighted by a concept of Federated (virtual) power plant~\citep{morstyn2018p2p}, where the P2P trading encourages prosumers to form the power plant -- through coalitions -- and consequently realising the prosumers values to power system value-chain.

Although the conducted research in both streams potentially provides economical benefits to involved parties, the \emph{heterogeneity} in prosumers' preferences~\citep{baarslag2017} -- attributed to personalized allocation for prosumers -- have not been considered in deciding the energy allocations and flexibility coordination. 
Additionally, without proper coordination of DERs and flexibilities, an energy community may face technical and regulatory challenges~\citep{Gautier2018}.  Moreover, the curtailment of feed-in due to excess supply of renewable energy (e.g. from rooftop PV) and consequent reverse power flow arising from uncoordinated and individually controlled DER maps to the \emph{loss of opportunity} phenomena that could otherwise be avoided by the synergetic and coordinated exchange of DER among prosumers. 
Only recently, several noteworthy research considered the \emph{heterogeneity} in prosumers' preferences, where the \emph{heterogeneity} arises from how a prosumer perceives different societal and environmental aspects such as energy contracts, generation technologies, locations of the network, owners' reputation, and so on in different parts of distribution networks~\citep{morstyn2019}, particularly through P2P contract network~\citep{Morstyn2018} and bilateral P2P negotiation strategy~\citep{chakra2018-1}. For instance, in~\citep{Morstyn2018}, the author presents an innovative bilateral contract network as a scalable market for P2P energy trading across the different value chain of electricity networks considering different types of players and their preferences. 

Automated P2P negotiation~\citep{baarslag2014bid} contrasts with the canonical P2P trading in that the former is an interactive decision making paradigm that provides a win-win outcome under partial/no information sharing environment, while the latter facilitates a more commodity-oriented trading platform - in more of an architectural paradigm - where the decision making is not imperative. Automated negotiation based iterative decision making is increasingly considered to be a promising facilitator of intelligent smart grid~\citep{alam2015, chakra2018-1}. 
For instance,~\citep{alam2015} proposed an automated negotiation protocol that has been applied to address the energy exchange between off-grid smart homes. However, the designed protocol imposes several key restrictions, in which only two \emph{exchange periods} over a day in which only equal amounts \emph{energy volume} can be exchanged, and thereby has limited applicability in real settings. 
In~\citep{chakra2018-1}, an automated negotiation strategy is utilised for settlement of energy contract in residential cooperatives considering the heterogeneity of prosumers' preferences. 
The agents in~\citep{chakra2018-1} implement a random peer selection strategy that, on several occasions (around 20\% of the total negotiation sessions), leads to failed negotiations as the agent may choose an incompatible peer. This paper advances the work~\citep{chakra2018-1} by proposing a learning-based intelligent peer selection strategy that increases the quality of the reached \emph{agreements} in terms of fairness, and the success rate of negotiation. In this paper, the \emph{negotiation domain} is enhanced to accommodate contracts that potentially is an \emph{agreement} that may be missing in a limited \emph{negotiation domain} explored the previous work. Additionally, this paper analyses the P2P interactions for drawing insights regarding emerging cluster of prosumers with compatible strategies.



\begin{figure}[h]
\centering
\includegraphics[scale=0.28]{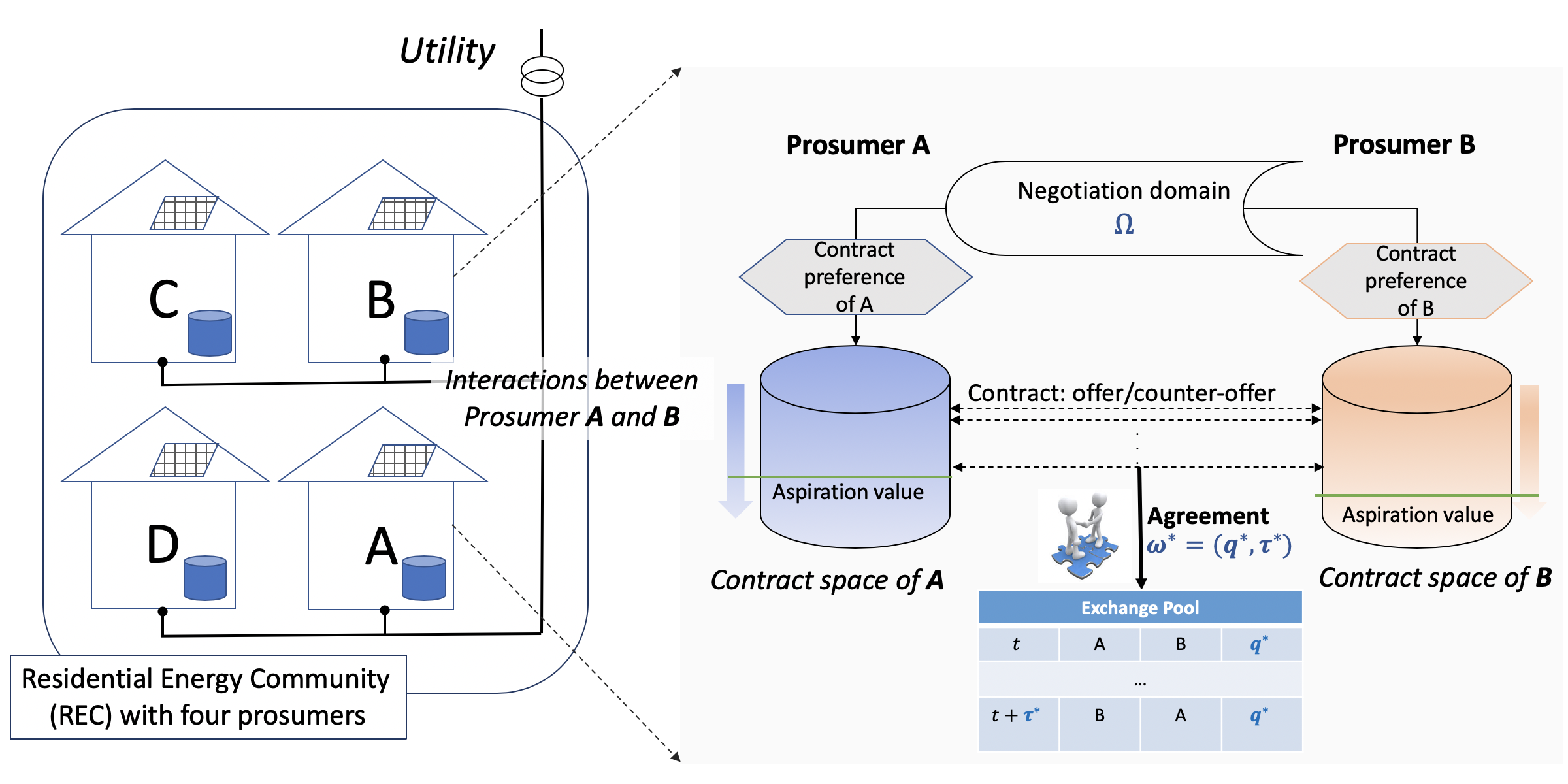}
\caption{An example of Residential Energy Cooperative (REC) with four prosumers/agents -- each comprising a PV panel and a residential battery -- are shown on the left side. On the right side, the interactions between two prosumers/agents (prosumer A and prosumer B) in settling energy contracts through P2P negotiation are zoomed-in and illustrated. While the detailed models are described in Section~\ref{system_model}, the associated concepts and definitions are put together in Section~\ref{sec_negotiation}.}
\label{fig_system_interactions}
\end{figure}

\section{Modelling Prosumers and Energy Contracts}
\label{system_model}
In this section, we present a prosumer by systematically modelling their load and generation profiles integrated with batteries. Later, we define \emph{energy contracts} with associated concepts followed by uncertainty modelling in planning that a potential \emph{energy contract} brings. 
The energy cooperatives are generally located at the Low Voltage (LV) distribution network where the prosumers are physically connected to exchange energy. However, we limit our contribution to the tertiary level of control without explicitly touching the physicality of the underlying distribution network such as power flow analysis. An outline of an REC with four prosumers, and the overall interactions between two agents are illustrated in Figure~\ref{fig_system_interactions}.

\subsection{Prosumer Model}
\label{sec_modelling_prosumer}
A prosumer is assumed to be equipped with a rooftop Solar Photovoltaic (PV) and a flexibility resource (e.g. battery energy storage system). We represent a prosumer as a software agent, $i \in N$, where $N$ is the set of agents in the cooperative. Let the (predicted) generation profile (through PV panels) of an agent $i$ be represented by $\widetilde{Pv}_{i}(t),~\forall t\in T$, where $T$ is the set of time-periods. 
Similarly, the predicted load profile of the agent $i$, at $t$ is represented by $\widetilde{Ld}_{i}(t)$. 
In addition, the battery dispatch (load) profile is denoted as $Pb_{i}(t)$, and a choice of agent $i$. The battery state of charge is modelled by the following equation
\begin{align}
X_i(t) = X_i(t-1) + \eta_b \times Pb_i(t) \times \Delta t - \epsilon_b,
\end{align}
where, $X_i(t)$ is state of charge (SOC) of the battery at $t$ and is operated within a limit. The constant degradation of the battery is represented by $\epsilon_b$. The dispatched battery power, $Pb_{i}(t)$ is constrained to operate within a limit. The efficiency of the battery, $\eta_b$ is dependent on whether the battery is being charged (with efficiency $\eta_{b}^{c}$) or discharged (with efficiency $\eta_{b}^{d}$)
\begin{align}
\eta_{b} = \left\{\begin{matrix}
 \eta_{b}^{c}, & \text{if}~Pb_i(t) \geq 0\\ 
 \frac{1}{\eta_{b}^{d}}, & \text{otherwise}
\end{matrix}\right.
\end{align}
After self-consumption, the net demand of agent $i$ becomes
\begin{align}
\widetilde{Ld}_i^{net}(t) = \widetilde{Ld}_i(t)-\widetilde{Pv}_i(t).
\end{align}
An agent $i$ engages in a trade with a subset of peers $j\in J \subseteq N$ at $t$, and the volume of energy being traded with each other is denoted as $ex_{i,j}(t)$.
The residual of agent $i$ -- after self-consumption, followed by the (cumulative) exchange with the peers and the local battery activation -- is the energy either wasted or to be traded on the external market, and is presented by the following energy balancing equation.
\begin{align}
\widetilde{Ld}_i^{res}(t) = \widetilde{Ld}_i^{net}(t)+ \sum_{j\in J}ex_{i,j}(t) + Pb_i(t).
\label{eq_net_load}
\end{align}

\subsection{Energy Contracts: P2P energy lending}
\label{sec_contract_def}
A simple but effective contract to exchange flexibility in energy systems is \emph{energy loan}~\citep{Claessen2016}, which we here adapt to the P2P setting.
In automated multi-issue negotiation, agents negotiate over several \emph{issues} with a target to achieve an agreement -- a value attribution to those issues -- that generates a socially optimal outcome for the participating agents. 
We consider energy loans parameterized by two important \emph{issues} over which the agents negotiate:
\begin{itemize}
\item[1.] The volume of energy to be traded between two agents (denoted by $q \in \mathcal{Q} \subseteq \mathbb{R}$, where $\mathcal{Q}$ is a discrete set of energy volumes).
\item[2.] The time of receiving the energy back (denoted by $\tau \in \mathcal{T} \subseteq \bm{Z^{+}}_{>0}$, where $\mathcal{T}$ is a discrete set of positive time periods).
\end{itemize}
A \emph{negotiation domain} $\Omega$ comprises all possible \emph{energy contracts}, i.e. $\Omega=\mathcal{Q}\times\mathcal{T}$. Every $\omega = (q,\tau) \in \Omega$ is a potential \emph{energy contract} (or loan) within the multi-issue negotiation that specifies a value for each \emph{issue}.

The energy volume $q$ and the return time $t+\tau$ influence respective energy profiles for both negotiating agents and consequently affect their local flexibility dispatch. For instance, executing an \emph{energy contract} $\omega = (q,\tau)$ between two agents $i$ and $j$ is reflected on the exchange vector $ex_{i,j}$, as follows:
\begin{align}
ex_{i,j}(t) &= q, \nonumber \\
ex_{i,j}(t+\tau) &= -q.
\label{eq_offer_impl}
\end{align}
According to the energy balancing equation (Eq.~\ref{eq_net_load}), a change in $ex_{i,j}$ changes the battery dispatch $Pb_i(t)$ while keeping the residual demand constant. The exchanged volume $ex_{i,j}$ is bounded by the physical link constraint that describes the maximum power to be exchanged between agents $i$ and $j$. Therefore,
\begin{align}
|ex_{i,j}| \leq z\times L_{i,j},
\label{eq_line_const}
\end{align}
where $L_{i,j}$ is the physical link limit between agents $i$ and $j$, and $z$ is the link efficiency.

As depicted before, agents may have varied preferences over a predefined set of criteria, i.e. the agents tend to weigh the criteria differently. Let $\mathbb{C}$ represent the set of criteria upon which the agents state their preferences. In this paper, we assume an agent evaluates an energy profile -- resulting from an \emph{energy contract} -- based on two criteria: \emph{loss in flexibility} and \emph{autarky} i.e. $\mathbb{C}:=\left\{~c_1=\textrm{loss in flexibility},~c_2=\textrm{autarky} \right\}$. 
The weight an agent $i$ places on criterion $c\in \mathbb{C}$ can be represented as a scaler $\lambda_{c}^{i}$ (where $\lambda_{c}^{i}$ are normalized weights, i.e. $\sum{_c}\lambda_{c}^{i}=1$) and is private to $i$. We assume, the weights are known to agent.
An \emph{evaluation function}, $e_{c,t}^{i}(\omega)$ is defined that denotes how an \emph{energy contract} performs, at $t$, from the perspective of criterion $c$ given the private preferences of agent $i$. 
Additionally, a planning horizon, $w$ defines how far ahead of an agent looks while deciding about the \emph{contracts}. The planning horizon depends on the uncertainty on the demand/generation prediction.

\textit{Criterion 1:} Criterion \emph{loss in flexibility} measures the emergent loss (in energy) due to the round-trip efficiency of the flexibility (e.g. battery) dispatch resulting from implementing an \emph{energy contract}. Particularly, activating (i.e charging/discharging) a battery incurs loss as the round-trip efficiency reduces the amount of energy. 
The \emph{evaluation function} associated with \emph{loss in flexibility} is defined as
\begin{align}
e_{c_1,t}^{i}(\omega)=\sum_{k=t}^{k=t+w}Pb_i(k) + \Theta(X_{i}(t:t+w)),
\label{eq_ec_1}
\end{align}
where $X_{i}(t:t+w)$ represents a vector of SOC profile of the battery from period $t$ to $t+w$ and the function $\Theta(.)$ calculates the offset power requires complete a full cycle of the battery, i.e. to bring the final SOC (at $t+w$) equals to initial SOC (at $t$).
Therefore, $\omega$ directly influences the battery dispatch power $Pb_i(t)$ through the energy balancing equation, i.e. Eq.~\ref{eq_net_load}.

\textit{Criterion 2:} Autarky in an \emph{energy contract} signifies the sustainability of the contract, which actually measures the total (estimated) energy to be traded on the external market provided that the \emph{energy contract} is implemented.
The \emph{evaluation function} associated \emph{autarky} is formally defined as
\begin{align}
e_{c_2,t}^{i}(\omega)=\sum_{k=t}^{k=t+w}|\widetilde{Ld}_i^{res}(k)|.
\label{eq_ec_2}
\end{align}
Agent aggregates the weighted \emph{evaluation function} of individual criterion to measure the quality of an \emph{energy contract}. The \emph{utility} function is defined as
\begin{align}
U_{i,t}(\omega)= \sum_{c}\lambda_{c}^{i} e_{c,t}^{i}(\omega).
\label{eq_utility}
\end{align}

\subsection{Uncertainty Modelling in Net Demand}
\label{subsec_uncert_model}
The load profile $\widetilde{Ld}_{i}(t)$ and generation profile $\widetilde{Pv}_{i}(t)$ of an agent $i$ are predicted signals and are potential sources of uncertainties. 
The \emph{utility} function defined in Eq.~\ref{eq_utility} depends on the point prediction of the net demand and therefore, may potentially inadequate of providing robust planning of local flexibility under uncertainty. 
To overcome this challenge, we utilise a set of stochastic scenarios of predicted net load profiles $\widetilde{Ld}_{i}^{net}$ and calculate the \emph{expected utility} of an \emph{energy contract}~\citep{chakra2016sg}. The scenarios of predicted net load profile are generated by taking samples from a \emph{Gaussian Process} comprising of 
\begin{itemize}
	\item a \emph{Gaussian} error Probability Density Functions (PDF), for each of the discrete lags $l$ in planning horizon $w$, i.e. $l=0,\cdots,w-1$, and 
	\item a \emph{Gaussian} PDF that models the interdependency between net load of two consecutive periods. 
\end{itemize}
A scenario $s \in \mathcal{S}$ of the predicted net load demand is calculated as 
\begin{align}
\widetilde{Ld}_{i}^{net}(t+l|t,s)=\widetilde{Ld}_{i}^{net}(t+l|t) + d_{i}(l,s),
\end{align}
where $d_{i}(l,s)$ is sampled from the aforementioned \emph{Gaussian Process}, and $\widetilde{Ld}_{i}^{net}(t+l|t)$ is the predicted net 
demand for period $t+l$ when predicted at $t$. Similarly, the residual net demand, defined in Eq.~\ref{eq_net_load}, for the scenario $s$ is restructured as
\begin{align}
\widetilde{Ld}_i^{res}(t+l|t, s) = \widetilde{Ld}_i^{net}(t+l|t, s)+ \sum_{j\in J}ex_{i,j}(t) + Pb_i(t+l|t, s).
\end{align}
Consequently, $e_{c_1,t}^{i}(\omega, s)$ and $e_{c_2,t}^{i}(\omega, s)$ are be redefined for the scenario $s$ - as shown below:
\begin{align}
e_{c_1,t}^{i}(\omega, s)&=\sum_{l=0}^{l=w-1}Pb_i(t+l|t, s) + \Theta(\bf{X_{i}}, s), \nonumber \\
e_{c_2,t}^{i}(\omega, s)&=\sum_{l=0}^{l=w-1}|\widetilde{Ld}_i^{res}(t+l|t, s)|.
\label{eq_criteria_scn}
\end{align}
Therefore, the \emph{utility} of an \emph{offer} $\omega$ for the scenario $s$ is modified as:
\begin{align}
U_{i,t}(\omega, s)= \sum_{c}\lambda_{c}^{i} e_{c,t}^{i}(\omega, s).
\end{align}
Finally, the \emph{expected utility} an \emph{energy contract} considering the uncertainties is defined as 
\begin{align}
\mathbb{E}U_{i,t}(\omega) &=\sum_{s\in \mathcal{S}}Pr(s) \cdot U_{i,t}(\omega,s),
\label{eq_ex_utility}
\end{align}
where $Pr(s)$ is the probability of the scenario $s$ and $U_{i,t}(\omega, s)$ is the modified \emph{utility} of an \emph{offer} $\omega$ considering 
the net predicted load scenario $s$. We assume the scenarios are equiprobable, and thus making $Pr(s)=1/|\mathcal{S}|$. Hereafter, we use the term \emph{utility} to represent \emph{expected utility} for the ease of description and to avoid potential confusions. 
The pseudo-predictions of the net demand are generated by adding a \emph{Gaussian} noise to the real net demand signal. 

\section{A Negotiation based Exchange Mechanism}
\label{sec_negotiation}
This section describes a negotiation based exchange mechanism with concept definitions of different contextual aspects of automated negotiation, as referred in Figure~\ref{fig_system_interactions}.
The algorithms describing the pairwise negotiation process and related procedures are presented as well in this section followed by a toy example to further clarify the mechanism. Finally, the quantification of efficiency and fairness of the proposed mechanism is detailed, which will be utilised to measure the performance of the proposed method.

Given the residential energy cooperative settings of several connected prosumers, the negotiation process may be understood as a \emph{multilateral negotiation}, emerging from multiple bilateral P2P pairwise negotiations\footnote{An alternative approach could be \emph{multi to multi negotiation}. We do not entertain that option as \emph{multi to multi negotiation} typically requires a mediator or a centralized coordination, as per the current state of the research.}. As the \emph{negotiation protocol}, we implement the \emph{alternating offers protocol}~\citep{baarslag2014bid}, which is commonly used in automated multi-issue negotiation settings.
At a particular time period, the protocol assumes that each agent only engages in once for a P2P negotiation. 

\subsection{Aspects of automated negotiation}
To fully grasp the applicability of automated negotiation in an \emph{energy contract} settlement, several important aspects of the negotiation process are defined in the following subsection. 

\begin{mydef}
\label{def_c_space}
\textbf{Contract space}: A negotiating agent maintains an ordered \emph{contract space} of potential offers comprising the set of the issues in the \emph{negotiation domain}, $\Omega$. 
The space is ordered according to the \emph{utilities} of the \emph{contracts}- defined in Eq.~\ref{eq_ex_utility}. 
\end{mydef}

\begin{mydef}
\textbf{Aspiration region}: The \emph{aspiration region} defines the area in the target utility space within which an agent aspires to strike a deal with the peers. The region is bordered by an aspiration value -- defined by the agent and is private to that agent -- that is specified by a quantile of the distribution of $\mathbb{E}U_{i,t}(\omega),~\forall\omega\in\Omega$. In a sense, the \emph{aspiration value} determines the \emph{degree of cooperativeness} of an agent; a higher quantile represents lower cooperation and vice-versa.
\end{mydef}

\begin{mydef}
\textbf{Agreement}: An \emph{agreement} is an \emph{energy contract} that is approved by both negotiating agents, and can be denoted by $\omega^{*}=(q^{*},\tau^{*})$.
\end{mydef}

\begin{mydef}
\label{def_deadline}
\textbf{Deadline}: The maximum number of rounds of a negotiation before which the negotiating agents should reach an \emph{agreement}. If no \emph{agreement} is formed after the \emph{deadline}, the negotiation fails, and the agents implement the perspective no-exchange deal, i.e. they walk-away with the \emph{reservation value}.
\end{mydef}

\begin{mydef}
\label{def_res_val}
\textbf{Reservation value}: The private value a negotiating agent keeps as an outside option in case of a disagreement. In this work, we define reservation value as the utility an agent perceives by contemplating no exchange with peers, i.e. $\mathbb{E}U_{i,t}(\omega)$, when $\omega:=\{q=0\}$.
\end{mydef}

\begin{mydef}
\label{no_deal_sl}
\textbf{No-deal solution}: The \emph{no-deal solution} resulting from the situation when the agents choose not to engage in negotiation and consequently do not exchange energy, i.e. $\omega=\{0.0, 0\}$ with peers.
\end{mydef}

\begin{mydef}
\label{def_nash}
\textbf{Fair outcome}: An important measure to quantify the fairness in an \emph{outcome} that could be determined by the \emph{Nash solution}. 
The \emph{Nash solution} is the outcome that maximizes the product of the utilities (Eq.~\ref{eq_ex_utility}) acquired from an \emph{energy contract} of negotiating agents (e.g. agent $i$ and $j$).
\begin{align}
\omega_{\rm{Nash}}(t)=\underset{\omega\in\Omega}{\max}~\mathbb{E}U_{i,t}(\omega)\cdot \mathbb{E}U_{i,t}(\omega).
\label{eq_nash_solution}
\end{align}
\end{mydef}

\begin{figure}[h]
\centering
\includegraphics[scale=0.6]{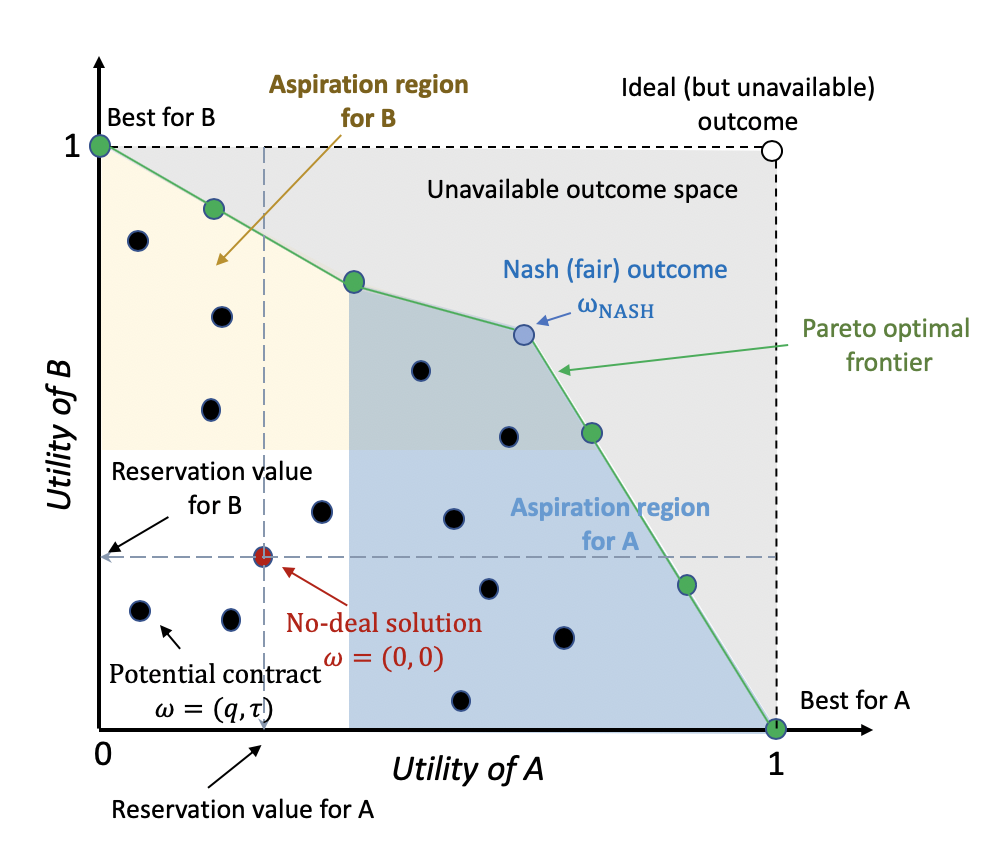}
\caption{Illustration of a \emph{contract space} of two agents in relation to the definitions. The black dots represent the potential contracts (or \emph{agreement} (i.e. $\omega$), while the \emph{Nash (Fair) outcome} is shown as a blue dot. The \emph{Pareto Frontier} hosts the set of contracts that are Pareto efficient\citep{moulin2004fair}. The agent specific \emph{aspiration regions} and \emph{reservation values} are highlighted}.
\label{fig_conceptual_outcome_space}
\end{figure}

\begin{figure}[h]
\centering
\includegraphics[scale=0.35]{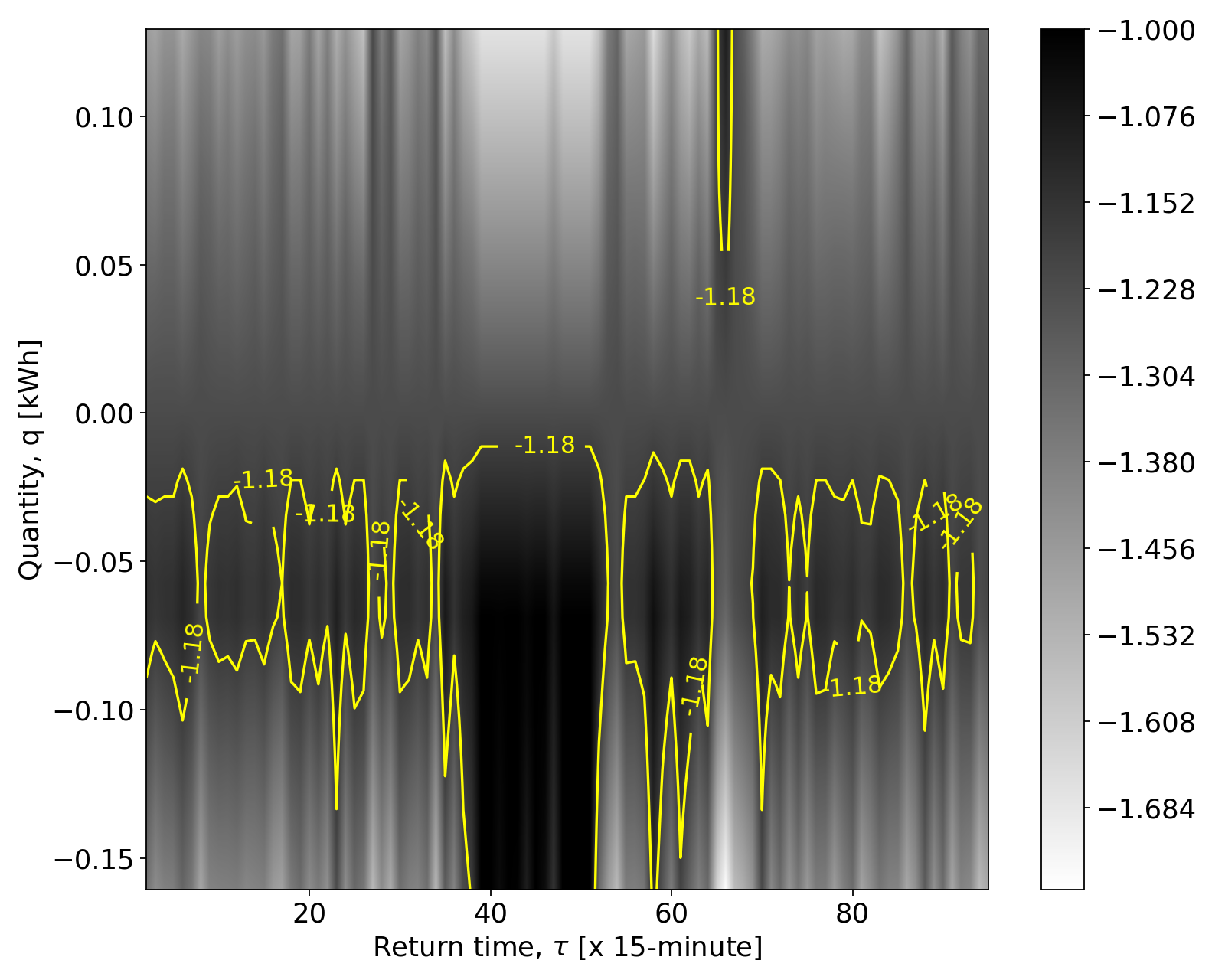}
\caption{An exemplary \emph{issue space} representing a \emph{negotiation domain} with acquired \emph{utility}. The \emph{issues} are presented at the axes whereas the \emph{utility} of the contracts is plotted as a heat-map. The contoured (darker) regions are the \emph{aspiration region} as they contain the preferred \emph{energy contracts} with the \emph{utility} higher than the \emph{reservation value}.}
\label{fig_strategy_space}
\end{figure}



\begin{figure}[h]
\centering
\includegraphics[scale=0.6]{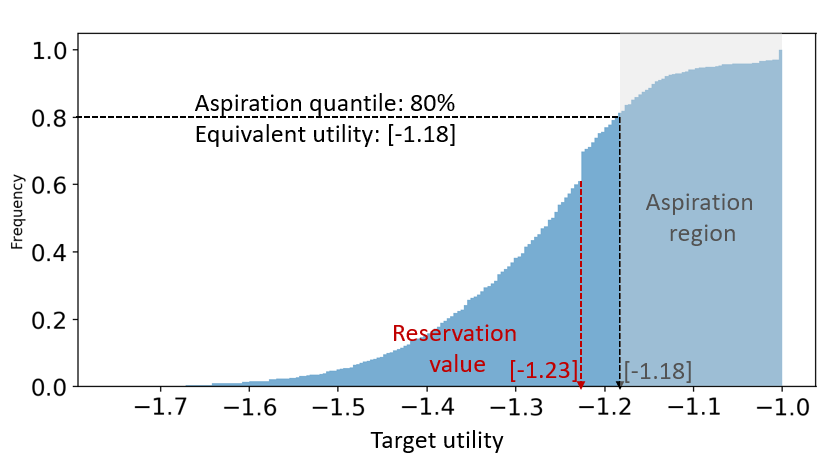}
\caption{Distribution of the \emph{target utility} of the agent described in Figure~\ref{fig_strategy_space}  with \emph{aspiration quantile} of 80\%. The \emph{aspiration region} is grey-ed out that results in the utility within $\textrm{[-1.0, -1.18]}$.}
\label{fig_utility_distribution}
\end{figure}

In a \emph{negotiation process}, the agents distributively (and interactively) search through the $\Omega$ to jointly agree on an \emph{energy contract} that maximizes their perspective \emph{expected utilities}. 
The problem of finding an \emph{agreement} and the definitions above are illustrated in Figure~\ref{fig_conceptual_outcome_space}. Agents' \emph{aspiration regions} are shaded in the Figure, whose intersection region represents the area where the agents prefer to strike a contract.
To further elucidate the negotiation process in context of the aforementioned definitions, Figure~\ref{fig_strategy_space} is presented that highlights an exemplary \emph{negotiation domain} and the associated \emph{utility} of all possible \emph{energy contracts} calculated from the perspective of an agent that has an \emph{aspiration quantile} of 80\% of the distribution that results in a value of $\textrm{-1.18}$, i.e. the agent aspires to receive at least $\textrm{-1.18}$ as the \emph{utility} from a contract. The demarcation of the \emph{aspiration regions} in the \emph{negotiation domain} are outlined through the contoured line. 
The figure illustrates that the agent prefers to provide energy to its peers - resulting in high utility region for negative quantities without caring so much about the return time, and it has a limited window of receiving energy from peers.
The distribution of the \emph{target utility} perceived by the agent for the same example is presented in Figure~\ref{fig_utility_distribution}. The \emph{contract space} is apparent from the figure - as the negotiation commenced, 
the agent starts bidding potential \emph{offers} that incur the higher \emph{target utilities} down to the \emph{aspiration quantile}, which is $\textrm{-1.18}$. 
By doing so, the agent follows a sophisticated \emph{bidding strategy} that traces the agent's utility distribution over potential offers within the \emph{aspiration region}.

\subsection{Peer selection process}
\label{sec_peer_selection}
The peer selection process is conducted by implementing an $\epsilon$-Greedy policy centred around the Nash (fair) solutions. An agent will be beneficial to pair-up with another agent where the mutually consented \emph{agreement} is more likely to be the Nash solution (Definition~\ref{def_nash}) for a particular negotiation session.
The iterative process implements an $\epsilon$-Greedy policy that chooses a peer -- drawing parallels from choosing an \emph{action} in a general \emph{Reinforcement Learning} framework-- that historically provides the fairer solution, with a probability of $(1-\epsilon)$, or a random peer with a small probability of $\epsilon$. Due to the implicit learning of what would the best peer for an agent, the agent gradually learns to make an intelligent peer selection call as negotiation process iterates over time. 
Implementing such a learning based peer selection process would increase the quality of achieved \emph{agreement} and reduce the likelihood of failed negotiation sessions as the agents are already equipped with historical information of pairwise outcomes with other agents.

\subsection{Algorithms: Pairwise negotiation process}
\begin{algorithm}[h]
\scriptsize
\DontPrintSemicolon
\Begin{
	$\Omega \longleftarrow \textrm{CREATE-Negotiation-Domain}(\mathcal{Q}, \mathcal{T})$\;
	$A.\textrm{GENERATE-Contract-Space}(\Omega, t)$\;
	$B.\textrm{GENERATE-Contract-Space}(\Omega, t)$\;
	$r \longleftarrow 0$\;
	$offerAccepted \longleftarrow False$\;
	\While{$r < deadline$~\rm{AND}~NOT~$offerAccepted$}{
	  \eIf{$r\%2=0$}{
	      $offer := A.\textrm{MAKE-Offer}(t)$\;
	      \If{$offer \neq \emptyset$}{
	      	$offerAccepted \longleftarrow B.\textrm{ACCEPT-Offer}(offer, t)$\;
	      	}
	    }
	    {
	      $offer := B.\textrm{MAKE-Offer}(t)$\;
	      \If{$offer \neq \emptyset$}{
	      	$offerAccepted \longleftarrow A.\textrm{ACCEPT-Offer}(offer, t)$\;
	      	}
	    }
	    $r\longleftarrow r+1$\;
	}
	\eIf{$offerAccepted$}{
	  $A.\textrm{IMPLEMENT-Agreed-Contract}(offer, t)$\;
	  $B.\textrm{IMPLEMENT-Agreed-Contract}(offer, t)$\;
	}{
	  $A.\textrm{IMPLEMENT-Reserve-Plan}(t)$\;
	  $B.\textrm{IMPLEMENT-Reserve-Plan}(t)$\;
	}

	return $offerAccepted$\;
}
\caption{NEGOTIATE-Contract($A$, $B$, $t$, $\mathcal{Q}$, $\mathcal{T}$)\label{al_negotiation}}
\end{algorithm}

\begin{algorithm}[h]
\scriptsize
\DontPrintSemicolon
\Begin{
	$agent.\textrm{Contract-Space} := \emptyset$\; \tcp*{set of offers; formation of the contract space}
	$agent.\textrm{Contract-Space-Hash} := \emptyset$\; \tcp*{hash for searching utility by offer}
	\For{$<q, \tau > \in \Omega$}{
		$offer := <q, \tau>$\;
		$utility \longleftarrow agent.\textrm{CALCULATE-Utility}(offer, t)$\;
		$agent.\textrm{Contract-Space} := agent.\textrm{Contract-Space} \cup \{offer, utility \}$\; \tcp*{populate the contract space}
		$agent.\textrm{Contract-Space-Hash}.set(offer) \longleftarrow utility$\; \tcp*{populate the hash}
	}
	$agent.\textrm{Contract-Space} := \textrm{ORDER-by-Utility}(agent.\textrm{Contract-Space})$\; \tcp*{order the contract space by utility}

}
\caption{agent.GENERATE-Contract-Space($\Omega, t$)\label{al_generate_contract_space}}
\end{algorithm}

\begin{algorithm}[h]
\scriptsize
\DontPrintSemicolon
\Begin{
	$net\_load := agent.\textrm{GET-Predicted-Net-Load}(t)$\; 
	$scenarios := agent.\textrm{GENERATE-Scenarios}(net\_load, t)$\; \tcp*{generate net-load scenarios according to Section~\ref{subsec_uncert_model}}

	$scenarios\_with\_offer := agent.\textrm{AMEND-Scenarios}(scenarios, offer, t)$\; \tcp*{amend the net-load scenarios with potential affect of the offer on them; based on Eq.~\ref{eq_offer_impl}}

	$eval\_criterion\_1 := agent.\textrm{EVALUATE-on-C1}(scenarios\_with\_offer)$\;  \tcp*{based on Eq.~\ref{eq_criteria_scn}; criterion \emph{loss in flexibility}} 
	$eval\_criterion\_2 := agent.\textrm{EVALUATE-on-C2}(scenarios\_with\_offer)$\;  \tcp*{based on Eq.~\ref{eq_criteria_scn}; criterion \emph{autarky}} 

	$utilities := agent.\lambda_1 \times eval\_criterion\_1 + agent.\lambda_2 \times eval\_criterion\_2$\; \tcp*{utility from each net-load scenarios; based on Eq.~\ref{eq_utility}}
	$exp\_utility \longleftarrow 0$\;

	\For{$utility \in utilities$}{
		$exp\_utility \longleftarrow exp\_utility + (1/|scenarios|)\times utility$\;	
	}

	return $exp\_utility$\;
}
\caption{agent.CALCULATE-Utility($offer$, $t$)\label{al_calculate_utility}}
\end{algorithm}

\begin{algorithm}[h]
\scriptsize
\DontPrintSemicolon
\Begin{
	$offer, utility := agent.\textrm{Contract-Space}.pop()$\; \tcp*{retrieve the current best offer with associated utility}
	$default\_utility \longleftarrow agent.\textrm{ASPIRATIONAL-utility}()$\;
	\If{$utility < default\_utility$}{
		$offer :=  \emptyset$\;
	}

	return $offer$\;
}
\caption{agent.MAKE-Offer($t$, $round$)\label{al_make_offer}}
\end{algorithm}

\begin{algorithm}[h]
\scriptsize
\DontPrintSemicolon
\Begin{
	$utility \longleftarrow agent.\textrm{Contract-Space-Hash}.get(offer)$\;
	$default\_utility \longleftarrow agent.\textrm{ASPIRATIONAL-utility}()$\;
	$response \longleftarrow False$\;

	\If{$utility > default\_utility$}{
		$response \longleftarrow True$
	}

	return $response$\;
}
\caption{agent.ACCEPT-Offer($offer$, $t$)\label{al_accept_offer}}
\end{algorithm}
In this section, we outline several key algorithms that are required to carry out a pairwise automated negotiation through \emph{alternating offer} protocol.
Algorithm~\ref{al_negotiation} describes
the high-level algorithm of the negotiation process between two agents $A$, and $B$ at time $t\in T$ via \emph{alternating protocol}. 
The Algorithm~\ref{al_negotiation} is presented in a centralized (i.e. non-agent) fashion for simplicity; an agent version of the same can be easily inferred from the algorithm.
The process starts with creating a negotiation domain $\Omega$ that will be used by both agents. 
Agents then generate perspective ordered \emph{contract space} by evaluating all possible contracts in $\Omega$ while considering their \emph{utility} over a planning horizon $w$. 
The detailed process of generating \emph{contract space} along with the \emph{utilities} of the contracts therein is described in Algorithm~\ref{al_generate_contract_space}. 
The calculation of \emph{utility} for an offer is outlined in Algorithm~\ref{al_calculate_utility}. 
Coming back to Algorithm~\ref{al_negotiation}, an \emph{alternating offers protocol} is implemented where, in each round, one of the agents proposes 
an offer (picked from the ordered \emph{contract space}) to the other agent until a \emph{agreement} is reached or the \emph{deadline} is encountered. 
The process of making and accepting an offer by an agent are detailed in Algorithm~\ref{al_make_offer} and \ref{al_accept_offer}, respectively.
In case an \emph{agreement} is reached, the agents implement the agreed \emph{energy contract}. Otherwise, the plans associated with the
\emph{reservation values} are implemented by each agent.
While implementing an \emph{energy contract}, an agent (for instance, $A$) amends to an existing exchange pool by stating \emph{how much} energy ($q^{*}$) 
to be traded with \emph{whom} (for instance, $B$) and \emph{when} ($t$) as well as by listing the \emph{same} volume of energy ($-q^{*}$) is committed to 
be traded back at $(t+\tau^{*})$ from $B$ (equivalent to Eq.~\ref{eq_offer_impl}).

As the negotiation and exchange proceed chronologically, the algorithm inherently exhibits a behaviour similar to that of a Model Predictive Control (MPC)~\citep{CHAKRABORTY2016405} methodology that controls the flexibility resources (e.g. battery). After a successful negotiation - shown in Algorithm~\ref{al_negotiation}, an agent implements the \emph{agreement} contract that effectively controls the battery with charge (or discharge) signals (as per power balance equation at Eq.~\ref{eq_net_load}). 
As we already know, for the \emph{agreement} that incurs \emph{utility} higher than the \emph{aspiration value} for perspective agents, the exchanged amount and the consequent flexibility controlling signals are already optimized and robust -- due to the inclusion of demand/generation uncertainty in decision making; Section~\ref{subsec_uncert_model} -- for the agents. 
Similar to MPC, an agent repeats the same process at each subsequent iteration while iteratively mitigating prediction errors for net demand.
Additionally, an agent may choose to deploy a sophisticated algorithm for controlling battery without harming the overall framework.

\begin{figure}[h]
\centering
\includegraphics[scale=0.53]{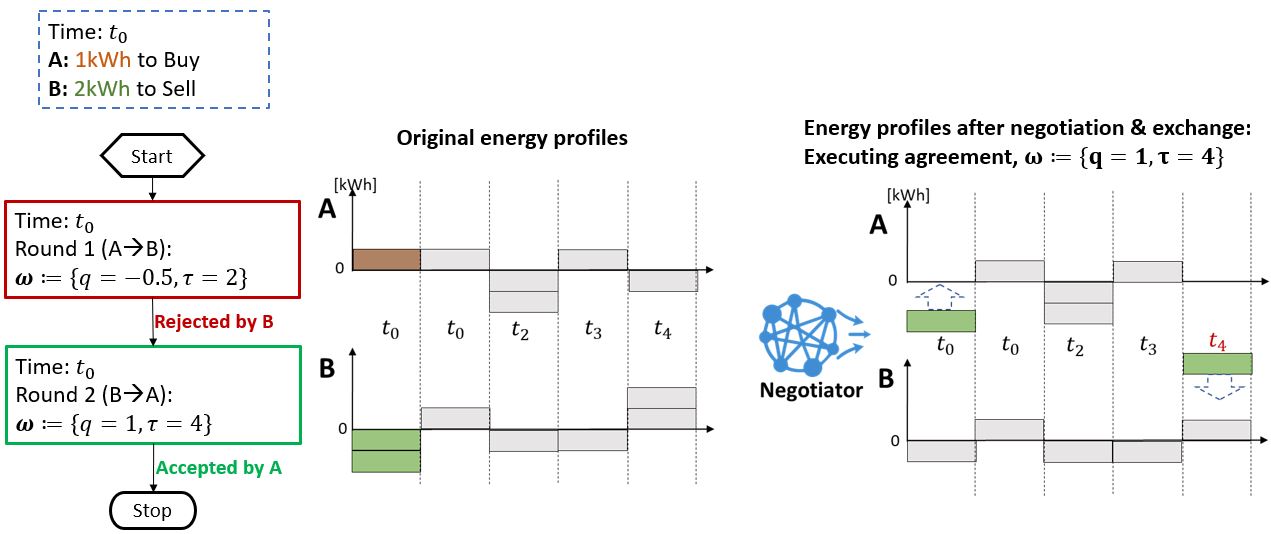}
\caption{Contract realization between two Agents $A$ and $B$.}
\label{fig_example}
\end{figure}

\subsection{Toy Example: Energy Contract}
Let's consider two agents (Agent $A$ and $B$) that are negotiating to settle for an \emph{agreement} with a planning horizon of 5 time-slots (e.g. $t_0 \cdots t_4$). The net demand profiles for Agent $A$ and $B$ are assumed as $\{ 1, 1, -2, 1, -1\}$  and $\{ -2, 1, -1, -1, 2\}$ kWh, respectively\footnote{In this example, we assume the temporal granularity to be 1 hour, so we can use kW and kWh interchangeably.}.
At $t_0$, $A$ requires 1 kWh while $B$ can provide 2 kWh of energy. The example is illustrated in Figure~\ref{fig_example}; the negotiation flow is shown in left while the effect of negotiation on agents are shown in right. 

The negotiation starts at round 1, when $A$ offers $B$ by asking $\textrm{-0.5}$ kWh at $t_0$ with a promise to give the energy back at $t_2$ i.e. the offer constitutes of $\omega:=\{q=-0.5, \tau=2\}$, according to Algorithm~\ref{al_make_offer}. 
$B$, after receiving the offer, evaluates the offer against its private preferences on the predefined criteria (Algorithm~\ref{al_accept_offer}), and rejects the offer as the incurred \emph{utility} from the offer falls below (assumed for the sake of the example) the \emph{aspiration value} of $B$.
The negotiation thus continues and at round 2, $B$ states the offer -- adhering to the \emph{alternating offer} protocol, Algorithm~\ref{al_negotiation} -- to $A$ as $\omega:=\{q=1, \tau=4\}$. Agent $A$ receives the offer and accepts it as the offer produces a \emph{utility} that falls within the \emph{aspiration region} of $A$. The negotiation thus stops with an \emph{agreement}. Note that, the offering agents at both rounds ($A$ at round 1 and $B$ at round 2) made offers from their higher (targeted) \emph{utility} spaces. Therefore, an agreement always stays in an intersection of \emph{aspiration regions} of participating agents.

\begin{table}[h]
\centering
\caption{Properties of the strategies.}
\label{tbl_scn}
\begin{tabular}{|r|r|r|r|}
\hline
Strategy & Local trading & Flexibility activation\\ \hline
\emph{No flexibility} & No & No \\ 
\emph{Individual control} & No & Yes  \\ 
\emph{Negotiation and control} & Yes & Yes \\ \hline
\end{tabular}
\end{table}

\subsection{Efficiency and Fairness}
\label{sec_eff_fair}
In addition to the proposed negotiation based strategy, we define a couple of complementary strategies to analyse the comparative efficiencies of the resulting allocations.
\begin{itemize}
\item \textit{No flexibility, $s_0$:} The prosumers do not activate their batteries and only trade residuals with external market.
\item \textit{Individual control, $s_1$:} This strategy is being currently utilised in the real residential setting, where the prosumers activate their local batteries, individually control the batteries and trade the residuals with external market. However, prosumers do not engage in trading with each other.
\item \textit{Negotiation and control, $s_2$:} The proposed strategy where prosumers engage in bilateral negotiation over \emph{energy contracts} with peers, implement the \emph{agreement}, and finally activate their batteries to control the residual energy. 
The remaining energy is traded in external market.
\end{itemize}

The properties of the strategies are briefed in Table~\ref{tbl_scn}. We define the \emph{performance} of an agent achieved by applying a particular strategy. 
The \emph{performance} of the proposed strategy $s_2$ considers the realized energy profile, after periodically negotiating and implementing the \emph{agreement}. The \emph{performance} is, therefore, similar to Eq.~\ref{eq_utility} except it takes into account the realized energy profile and the consequent battery dispatch.
\begin{align}
\xi_{i}(s_2)  = \lambda_{1}\times \left [ \sum_{t}^{T}Pb_i(t) + \Theta(\bf{X_{i}})\right] + \lambda_{2}\times \left [ \sum_{t}^{T}\left |Ld_i^{res}(t)\right | \right].
\end{align}
For strategy $s_{0}$, the $\xi_{i}(s_0)$ only considers the \emph{autarky} components - without the energy exchange component, i.e. $\sum_{j\in J}ex_{i,j}(t)$ in Eq.~\ref{eq_net_load}. And for strategy $s_{1}$, the $\xi_{i}(s_1)$ considers both criteria, but again without the flexibility component (i.e. $Pb_{i}(t) = 0$).

In order to validate the efficiency of the strategies $\mathbb{S}:=\{s_0, s_1, s_2\}$ in improving the social welfare, we define the \emph{utilitarian social welfare} as
\begin{align}
sw_{s} = \sum_i^{N} \xi_{i}(s),
\label{eq_sw}
\end{align}
for all $s\in \mathbb{S}$. Moreover, we quantify the relative fairness of a strategy $s$ (to another strategy $h$) based on the \emph{Nash social welfare} criterion, an established concept of fairness~\citep{moulin2004fair}, as following
\begin{align}
nw_{s|h} = \prod_i^{N} \left ( \xi_{i}(s) - \xi_{i}(h)\right ).
\label{eq_fairness}
\end{align}

\section{Numerical Simulation and Discussion}
\label{sec_sim}
In this section, we consider two cases of varied scaled cooperatives to empirically evaluate different aspects of the proposed strategy. 
\begin{itemize}
\item Case 1: \textit{Cooperative of 2 agents} presents the effects of the proposed strategy on the residual demand and consequent battery dispatch, and the agents' \emph{negotiation} domain exploring phenomena.
\item Case 2: \textit{Cooperative of 9 agents} verifies the quality of the allocation achieved by the proposed strategy from the perspectives of efficiency and fairness and investigates agents' P2P interactions.
\item Case 3: \textit{Cooperative of 100 agents} verifies the scalability of the proposed strategy.

\end{itemize} 
The aforementioned cases assume the local flexibility (i.e. battery) is owned privately and controlled individually by the prosumers. The proposed algorithms are implemented and the simulation is conducted using Python (version 3.7) programming language on a Windows machine (Intel Core i5 2.3 GHz with 16GB RAM). The total simulation period is taken as 7 days with 15-minute granularity, i.e. $\Delta t=15$. The planning horizon $w$ is set out to be 24-hours, which means an agent evaluates an \emph{energy contract} considering the potential effect the contract will have on its energy profile in the next 24 hours. The number of net-demand scenarios $|\mathcal{S}|$ (in Eq.~\ref{eq_ex_utility}) is set to 100. As for constructing the \emph{negotiation domain} $\Omega=\mathcal{Q}\times\mathcal{T}$, the set $\mathcal{Q}$ contains 15 discrete energy quantities, and the set $\mathcal{T}$ contains discrete time steps of $\{2, 3, \cdots, w\times\Delta t\}$. The \emph{deadline} for a negotiation session is set out to be 5000 rounds. The parameters are summarised in Table~\ref{tbl_sim_parameters}. The case-specific parameter settings are listed in Table~\ref{tbl_agent_spec} and Table~\ref{tbl_agent_spec_case_2} for Case 1 and 2, respectively. 

\begin{table}[t]
\centering
\caption{Parameters used for simulations}
\label{tbl_sim_parameters}
\begin{tabular}{|r|r|}
\hline
Total simulation period, $T$ & 7 days \\ \hline
Time granularity, $\Delta t$ & 15 minutes \\ \hline
Planning horizon, $w$ & 24 hours \\ \hline
Net demand profile scenarios, $|\mathcal{S}|$ & 100 \\ \hline
Number of discrete energy quantities, $|\mathcal{Q}|$ & 15 \\ \hline
Number of discrete time steps, $|\mathcal{T}|$ & $\{2, 3, \cdots, w\times\Delta t\}$ \\ \hline
Maximum rounds of negotiation session before deadline& 5000 \\ \hline
\end{tabular}
\end{table}

\begin{table}[t]
\centering
\caption{Agent Specifications: Case 1}
\scriptsize
\label{tbl_agent_spec}
\begin{tabular}{|r|r|r|r|r|r|}
\hline
Agent & Reservation (\%) & $\lambda_{c_1}$ & $\lambda_{c_2}$ & Capacity[kWh] & Efficiency \\ \hline
\bf{A} & 52 & 0.33 & 0.67 & 6.8 & 0.9 \\ 
\bf{B} & 50 & 0.71 & 0.29 & 7.0 & 0.8 \\  \hline
\end{tabular}
\end{table}
\normalsize

\begin{figure}[h]
\centering
\includegraphics[scale=0.35]{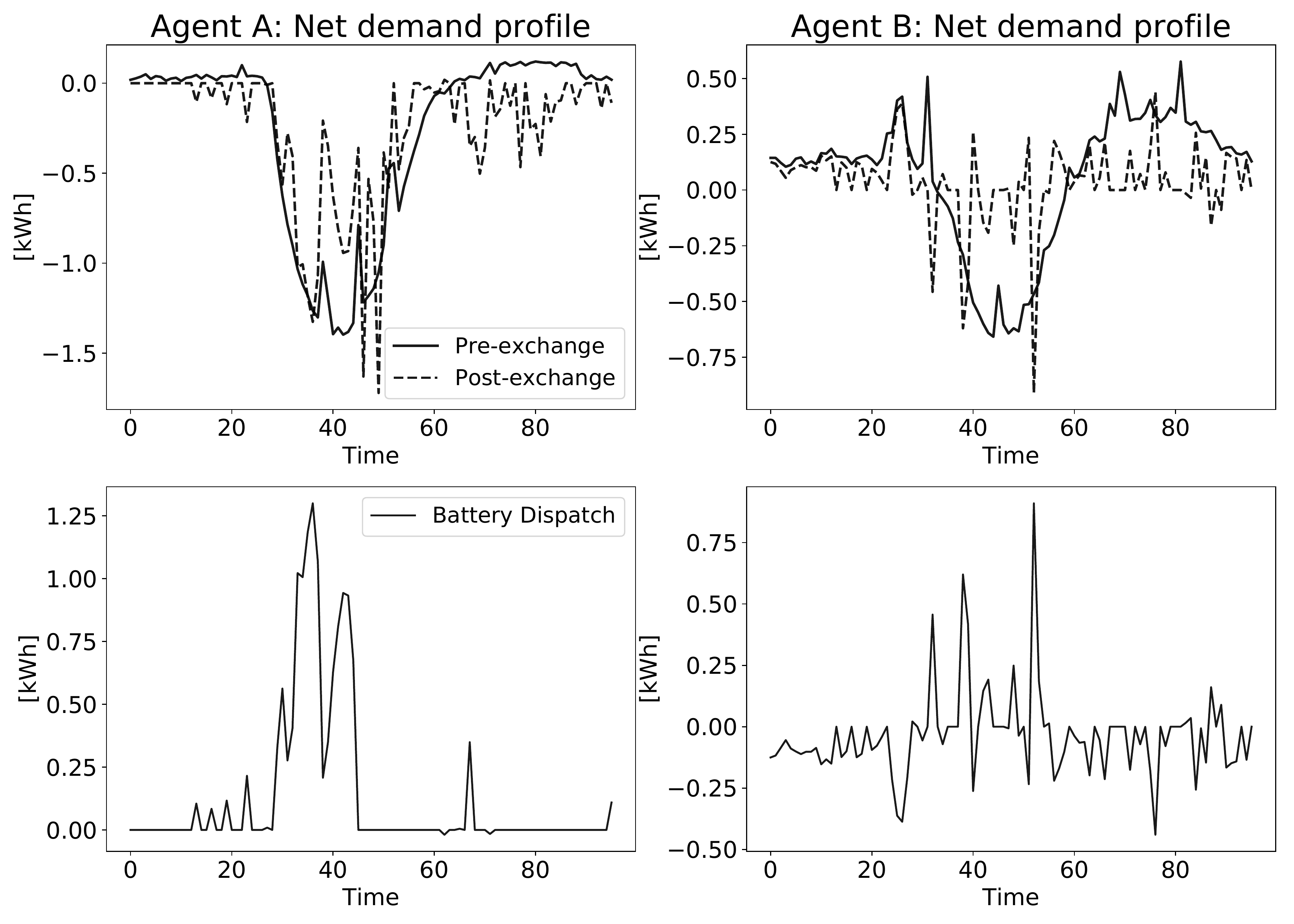}
\caption{Execution of an \emph{agreement}. Effects of executing \emph{agreements} -- on net demand and battery dispatch profiles -- reached through negotiation between two agents. }
\label{fig_outcome_space}
\end{figure}


\begin{figure}[h]
\centering
\includegraphics[scale=0.60]{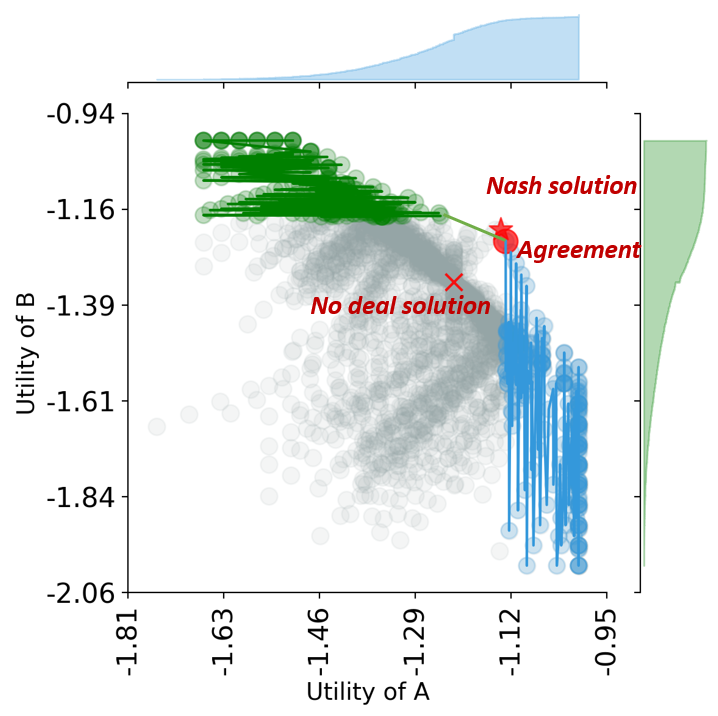}
\caption{Outcome space: The \emph{Agreement} is reached at the 1560-th round of negotiation sessions between the agents. An exemplary outcome space of \emph{agent A} and \emph{agent B} with marginal cumulative distributions of their \emph{utilities} are shown. The \emph{Nash solution}, \emph{agreement} and \emph{No-deal solution} are plotted to illustrate their relative distances and agents' capability to find a deal very close to the \emph{Nash solution} under the proposed negotiation based strategy.}
\label{fig_outcome_space_contour}
\end{figure}

\begin{figure}[h]
\centering
\includegraphics[scale=0.55]{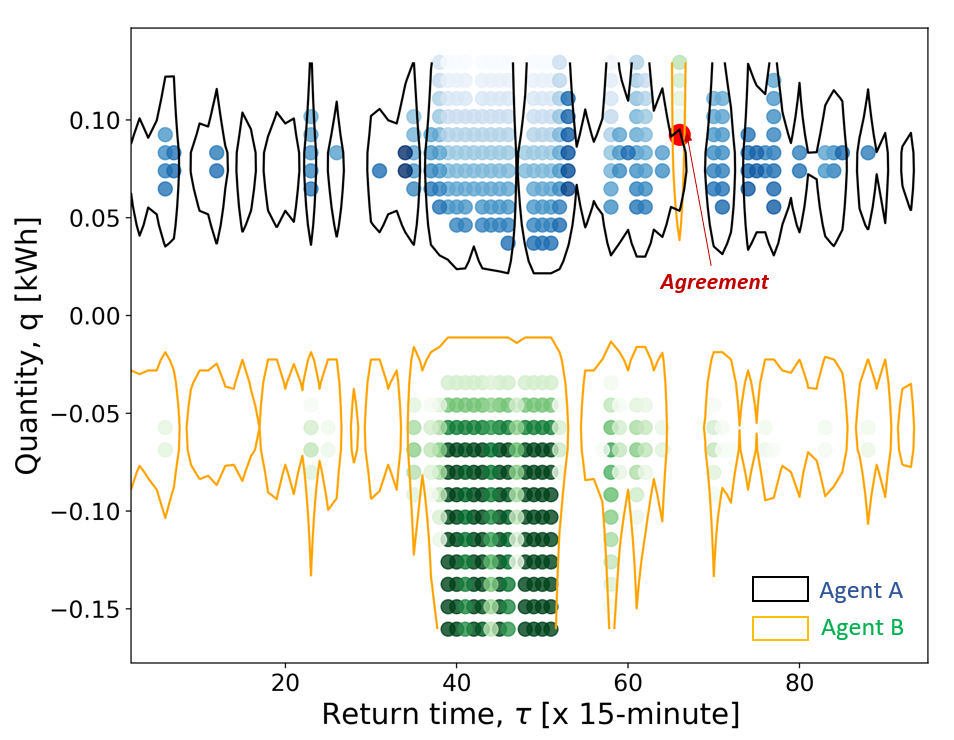}
\caption{Issue space: The \emph{negotiation domain} with the juxtaposed \emph{aspiration regions} bounded by contour lines for both agents. The negotiation traces, represented by the coloured scatters, depict the behaviour of the agents with the opponents as both of them converge to the \emph{agreement}. The gradients in the traces represent the trace order - the darker ones are with the higher \emph{utility}, which the perspective agents follow to reach the \emph{agreement}.}
\label{fig_contour_outcome_3}
\end{figure}

\subsection{Case 1: Cooperative of 2 Agents}
Flexibility activation through battery enhances the potential benefits as two agents could negotiate even when their net demand status are equal (i.e. both positive or negative).
The specification of the agents with associated battery information is provided in Table~\ref{tbl_agent_spec}. The charging and discharging rates of these batteries are 1.3kW and 3.3kW, respectively. The SOC of the batteries are operated within 20\% to 90\% of the respective 
capacities, and the degradation rate is set as 0.04\% of the same. As pointed at the Table, \emph{agent A} values criterion $c_2$ than 
criterion $c_1$; that is the agent places higher preferences on \emph{autarky}, while \emph{agent B} prefers \emph{loss in flexibility} more.

Figure~\ref{fig_outcome_space} shows effects of energy exchange (through negotiation) and resulting battery dispatch between \emph{agent A} and \emph{agent B}. The residual demand profiles resulting from negotiation clearly reflect the preferences of the agents. For instance, the battery dispatch profile of \emph{agent B}, who cares more about the \emph{loss in flexibility}, exhibits a relative fluctuating signal that results in an almost neutralized losses. 
The apparent fluctuations in the battery dispatches are due to the fact that they both agents implement a naive battery scheduling technique, as described in Section~\ref{sec_modelling_prosumer}. However, in the proposed framework, agents can easily mitigate such fluctuations by integrating an additional cost function - that penalizes such behaviour, into their utility 
function and subsequently placing a higher weight on that cost function.

Now, we analyse the exploration of \emph{negotiation domain} by agents as they reach \emph{agreements} by scrutinizing the \emph{issue space} and the \emph{outcome space}. 
For this experiment, the \emph{aspiration quantile} of the agents are kept identical, 80\% each, while the weights of the criteria are varied - \emph{agent A}: $\lambda_1=\textrm{0.33},~\lambda_2=\textrm{0.67}$, \emph{agent B}: $\lambda_1=\textrm{0.67},~\lambda_2=\textrm{0.33}$. Additionally, the round-trip efficiencies and the degradation rate of the agents' battery are diversified as well - round-trip efficiencies of \emph{agent A} and \emph{agent B} are set as 80\%, and the degradation rates are set as 0.04\% and 0.02\% (of battery capacity), respectively.
Figure~\ref{fig_outcome_space_contour} depicts a two-dimensional outcome space that emerges from the negotiation interactions between the agents and their marginal cumulative distributions of \emph{utility} over \emph{negotiation domain}. The trace of negotiation -- from the perspective of individual agents -- illustrate the power of heterogeneous preferences and the multi-issue setting, because the agents are able to exhaustively explore their \emph{iso-utility} curves and concede until an \emph{agreement} is found.  

Although the \emph{agreement} does not exactly reach the \emph{Nash solution}, it still yields \emph{utilities} that are located over 80\% quantile range of the distributions. As shown in the figure, the \emph{no-deal solution} (defined in Definition~\ref{no_deal_sl}) that attributes the situation when the agents do not engage in negotiation and consequently do not exchange energy, is located further down from the \emph{agreement}. Therefore, the \emph{agreement} and consequent energy exchange are more attractive for engaging agents. In addition, the \emph{agreement} is almost co-located with the \emph{Nash solution} and thus implying the fact that the agents almost managed to crack the fairest \emph{agreement}, which could not happen due to agents' individual preferences.
The issue space, represented by \emph{Return time} ($\tau$) vs. \emph{Quantity} ($q$), under the corresponding \emph{negotiation domain}, is illustrated in Figure~\ref{fig_contour_outcome_3}. The figure further points out the \emph{aspiration regions} of both agents and the phenomenon of how the \emph{agreement} is found at one of the intersections of the contoured \emph{aspiration regions}. As shown in the Figure, the individual traces -- represented by colour gradients -- of the \emph{utility space} exploration by each agent converge at the \emph{agreement}.


\begin{table}[t]
\centering
\scriptsize
\caption{Agent Specifications: Case 2}
\label{tbl_agent_spec_case_2}
\begin{tabular}{|r|r|r|r|r|r|}
\hline
Agent & Reservation (\%) & $\lambda_{c_1}$ & $\lambda_{c_2}$ & Efficiency (\%)& Degradation (\%)\\ \hline
\bf{1}	& 70 &	0.33 & 0.67	& 0.90	& 0.04 \\
\bf{2}	& 70 &	0.67 & 0.33	& 0.90	& 0.04 \\
\bf{3}	& 60 &	0.33 & 0.67	& 0.85	& 0.02 \\
\bf{4}	& 60 &	0.67 & 0.33	& 0.80	& 0.02 \\
\bf{5}	& 75 &	0.33 & 0.67	& 0.90	& 0.02 \\
\bf{6}	& 75 &	0.67 & 0.33	& 0.90	& 0.04 \\
\bf{7}	& 80 &	0.33 & 0.67	& 0.85	& 0.04 \\
\bf{8}	& 80 &	0.67 & 0.33	& 0.80	& 0.02 \\
\bf{9}	& 85 &	0.33 & 0.67	& 0.75	& 0.02 \\ \hline
\end{tabular}
\end{table}
\normalsize

\begin{figure}[t]
\centering
\includegraphics[scale=0.38]{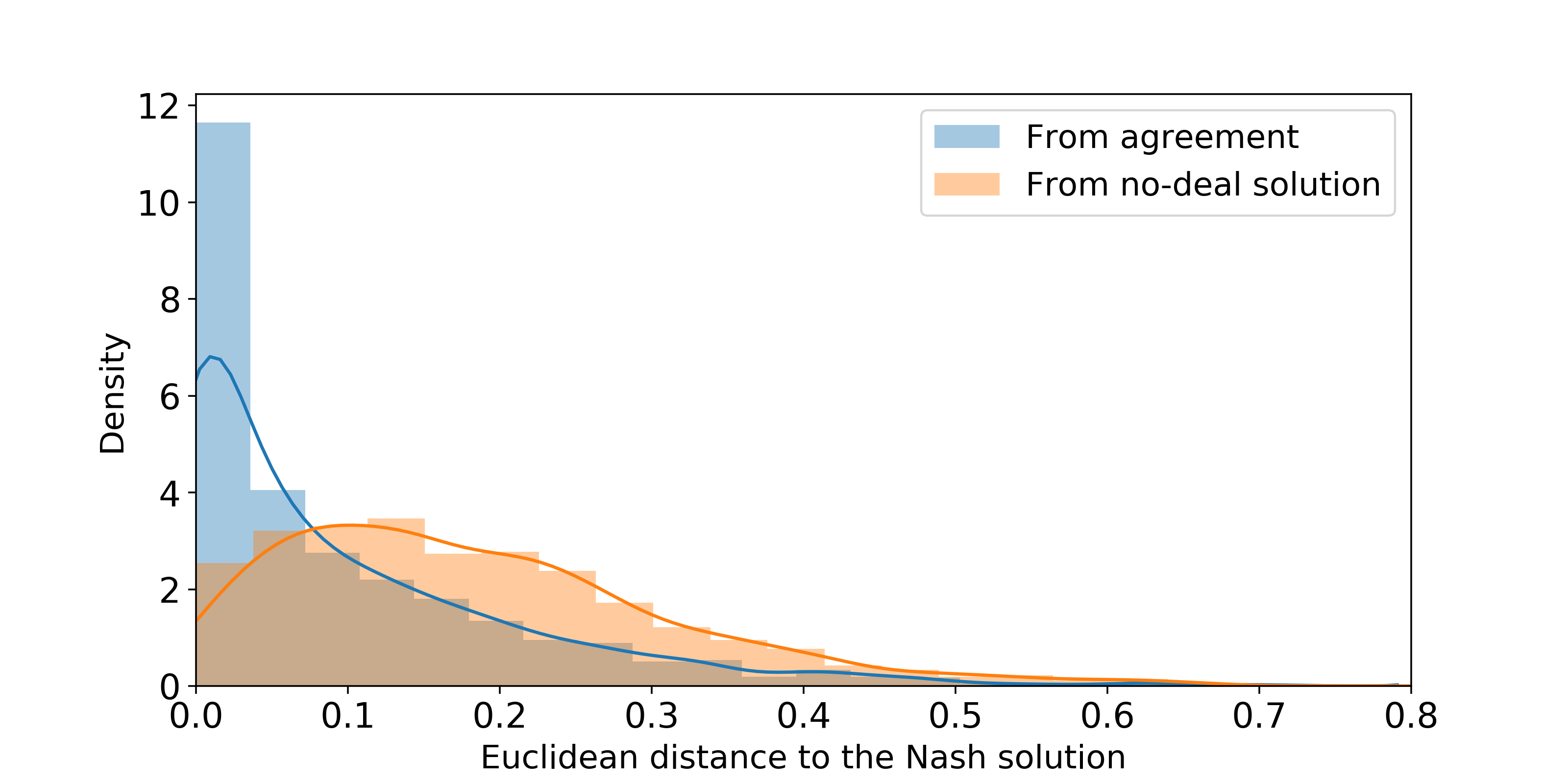}
\caption{Distributions of the 
Euclidean distance to the \emph{Nash solution} from the \emph{agreements} and the \emph{no-deal solutions} in the \emph{outcome space} and estimations of corresponding Gaussian kernel densities. The majority resulting \emph{agreements} are locating themselves closer to the \emph{Nash solutions}, hence confirm the fairness of the energy allocations.}
\label{fig_nash_distace}
\end{figure}

\begin{figure}[h]
\centering
\includegraphics[scale=0.4]{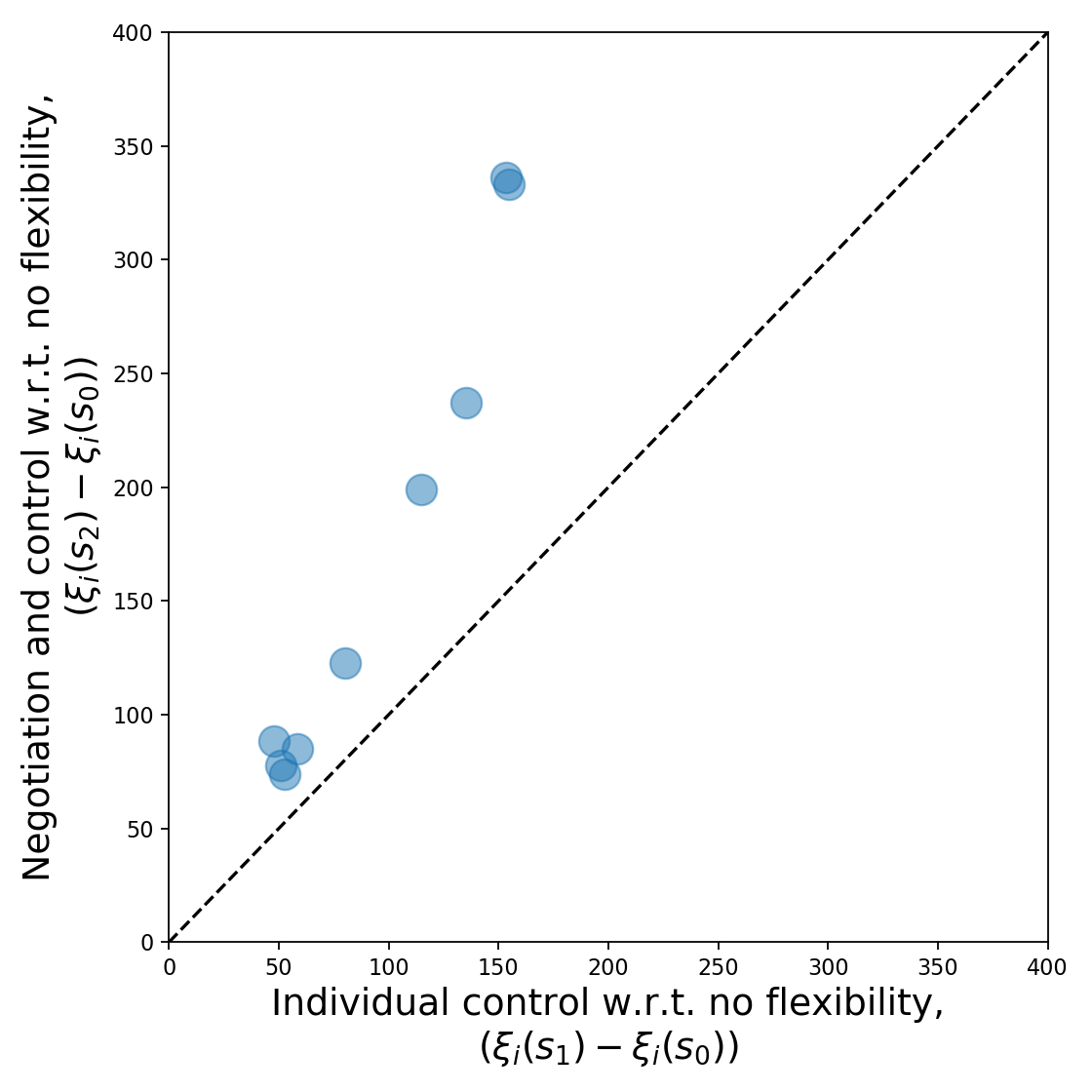}
\caption{Agents utility improvement of \emph{individual control} strategy (horizontal) and \emph{negotiation and control} strategy (vertical) over \emph{no flexibility} baseline. Agents are above the dashed equal improvements line, hence our newly proposed strategy dominates \emph{individual control}, while also improving relative fairness ($nw_{s_2|s_0} \approx 1.38 \cdot 10^{19} > nw_{s_1|s_0} \approx 3.35 \cdot 10^{17}$).}
\label{fig_improvement_sw}
\end{figure}

\begin{figure}[h]
\centering
\includegraphics[scale=0.60]{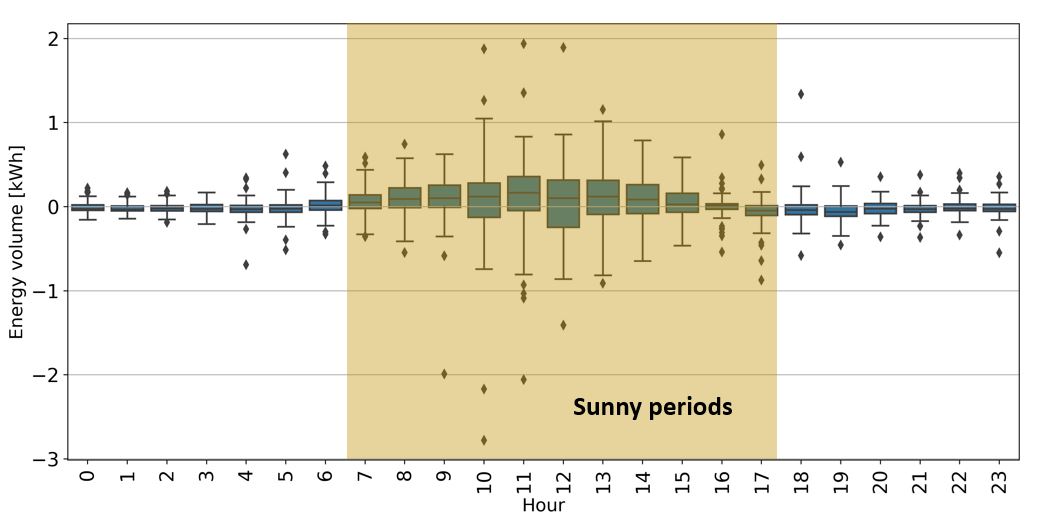}
\caption{Hourly distribution of the (agreed) energy volumes to be exchanged ($q*$) compiled over 7 days. Evidently, the agents are more flexible around the sunny hours and tend to exchange a larger volume of energy during these periods.}
\label{fig_q_dist}
\end{figure}

\subsection{Case 2: Cooperative of 9 Agents}

In this case, we analyse a higher scaled cooperative with 9 agents. The specifications of the agents are detailed in Table~\ref{tbl_agent_spec_case_2}. Overall, the experiments is run for a 7-day period (eq. 672 timestamps) with a success rate of 85\%, i.e. 15\% of the time the agents failed to strike any deal, and no exchange is performed. The success rate of the negotiation is improved by 5\% due to the deployment of an intelligent peer selection process (described in Section~\ref{sec_peer_selection}); the randomized peer selection strategy yielded to a success rate of 80\%.

\subsubsection{Allocation efficiency and fairness}
Figure~\ref{fig_nash_distace} elucidates the quality of the outcome through the distribution of the Euclidean distance from the \emph{agreement} to the \emph{Nash solution}, and how an \emph{agreement} outperforms a \emph{no-deal solution} by being more likely to be the \emph{Nash solution} for a negotiation session. 
The histograms are fitted through \emph{Gaussian distribution}, and it is apparent from the kernel density of the \emph{distance distribution} of the \emph{agreements} that the resulting allocations of energy derived from the negotiation based strategy are fairer than that of a non-negotiation based strategy. 

\begin{figure}[h]
\centering
\includegraphics[scale=0.32]{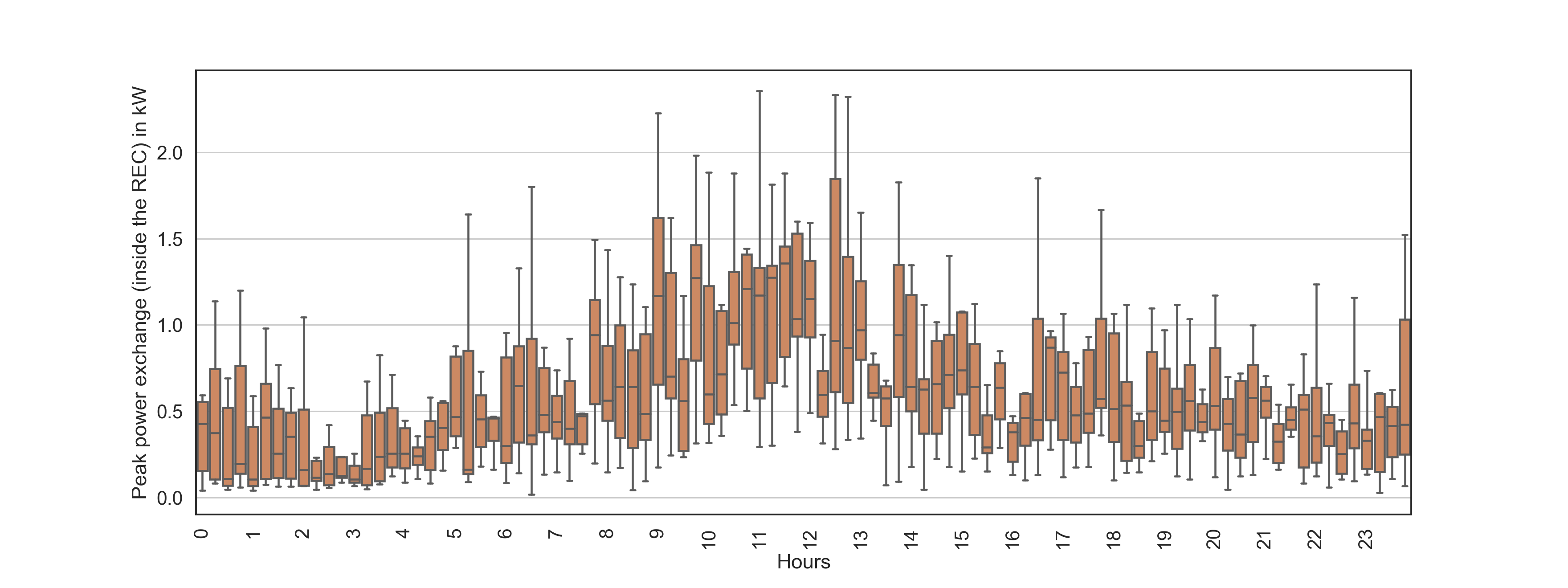}
\caption{Distribution of (absolute) peak exchange -- inside the REC and aggregated over all P2P exchange in every 15 minutes -- over 7 days of simulation period. The P2P negotiation and resulting exchange does not congest the physical network excessively. Relatively higher ramp of congestions could be noticeable during the sunny periods.}
\label{fig_total_trade_time}
\end{figure}

\begin{figure}[h]
\centering
\includegraphics[scale=0.48]{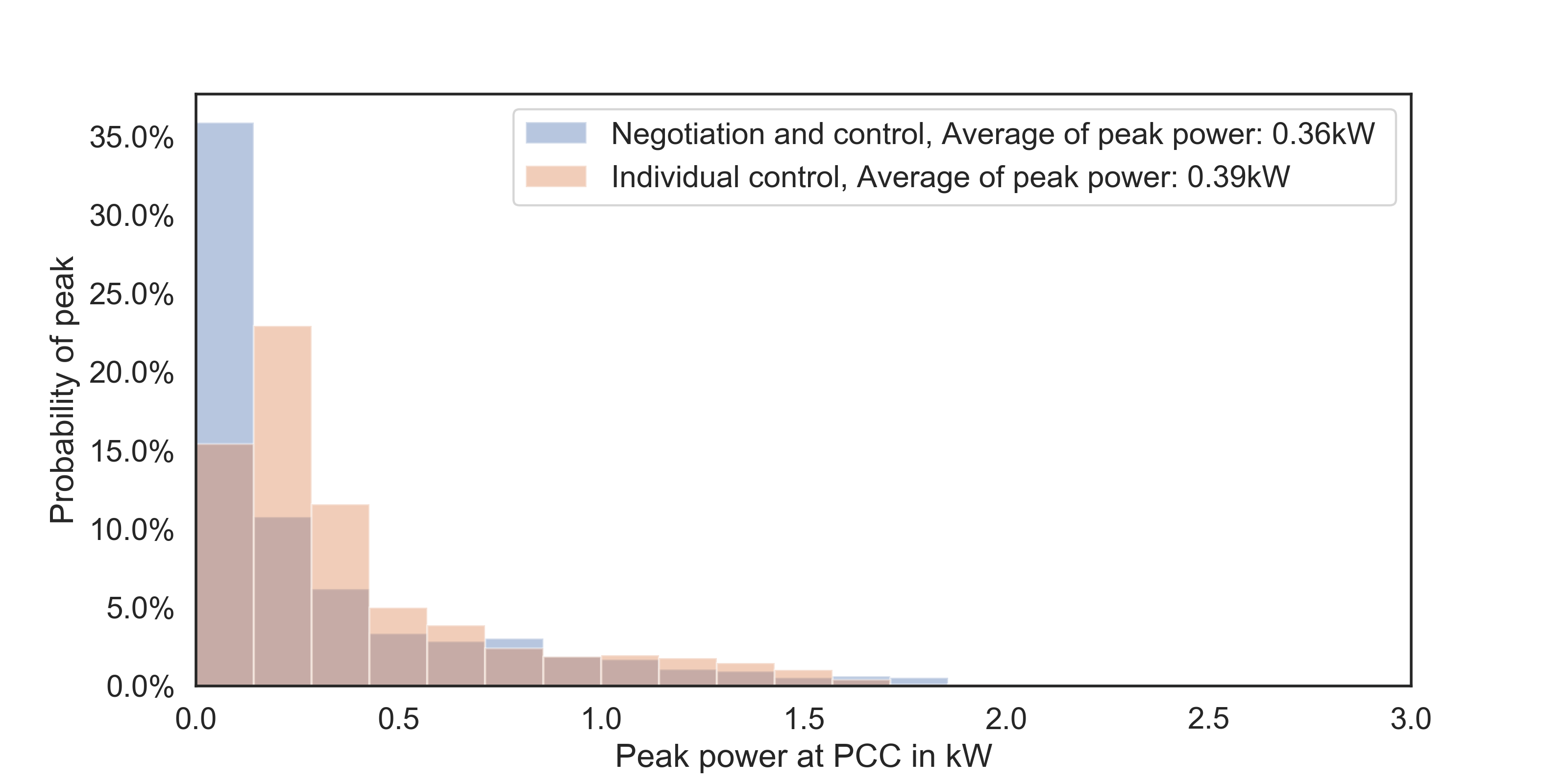}
\caption{Comparison of peak power distribution at PCC between the proposed \emph{Negotiation and control} strategy and \emph{Individual control} strategy. The representative distribution of \emph{Negotiation and control} strategy considers the return of the \emph{energy loan} as well. The resulting exchanges do not excessively congest the physical network in terms of peak power compared to that of \emph{Individual control}.}
\label{fig_total_trade_comp}
\end{figure}

\begin{figure}[h]
\centering
\includegraphics[scale=0.4]{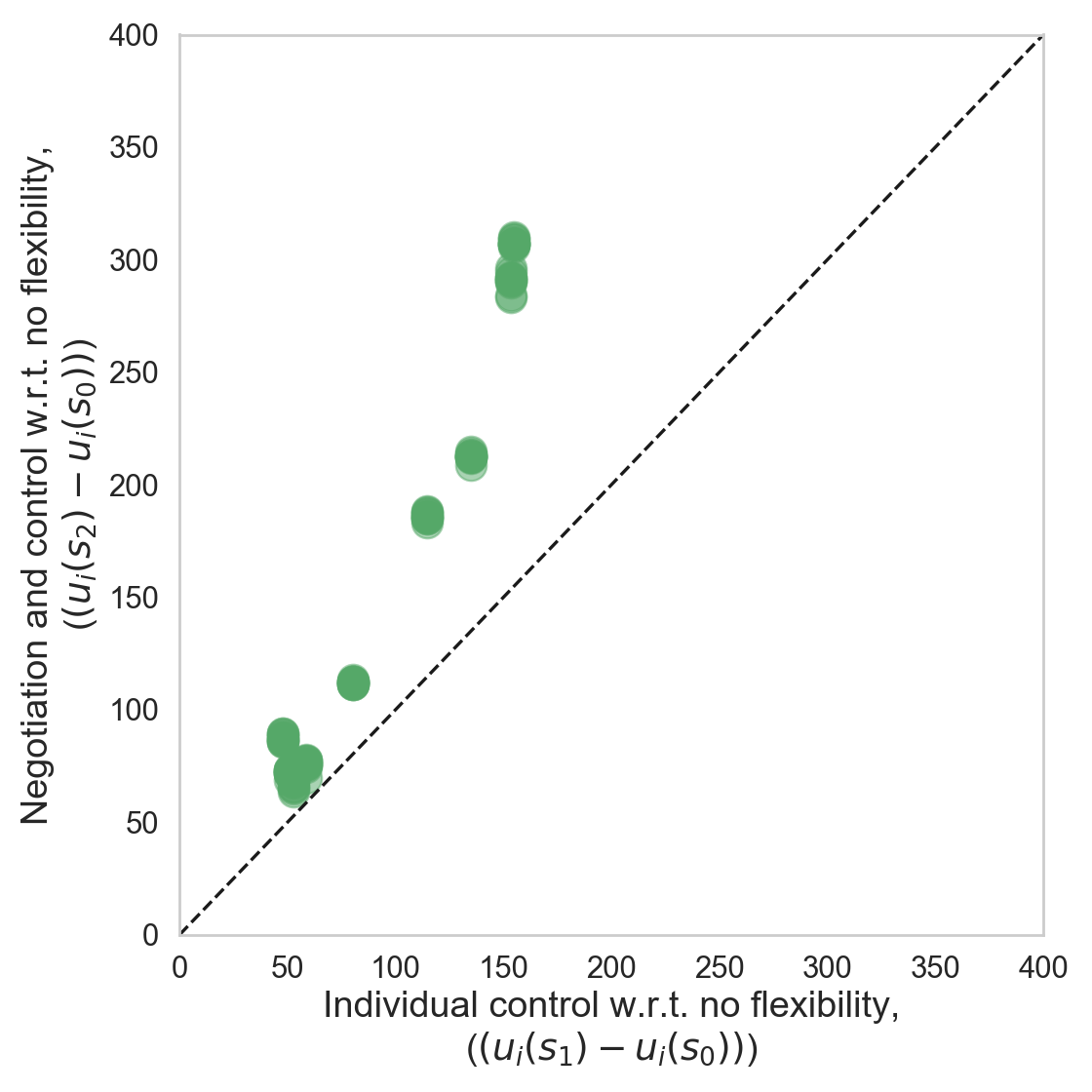}
\caption{Improvements in agents utility over the comparative strategies in an REC of 100 prosumers.}
\label{fig_improvement_sw_100}
\end{figure}

To further illustrate the performance of the proposed negotiation based strategy, we turn the analysis toward the allocative efficiency of the same, and discuss how the strategy establishes itself preferable for all agents over a baseline strategy of \emph{no flexibility} and a strategy of \emph{individual control} of flexibility without any P2P exchange, which were defined in Section~\ref{sec_eff_fair}. 
Figure~\ref{fig_improvement_sw} presents the relative increase in utility for each prosumer of an EC with 9 agents, comparing the improvements of \emph{individual control} strategy and \emph{negotiation and control} strategy over \emph{no flexibility} strategy. As seen in the figure, $(\xi_{i}(s_2)-\xi_{i}(s_0))$ dominates over $(\xi_{i}(s_1)-\xi_{i}(s_0))$ by placing itself over the equal-improvement dashed line. Therefore, it implies that the \emph{utilitarian social welfare} criterion is maximized by the proposed \emph{negotiation and control} strategy for all agents. Additionally, the relative fairness (measured by the \emph{Nash social welfare}, Eq.~\ref{eq_fairness}) is also improved by the proposed strategy, more specifically, $nw_{s_2|s_0} \approx 2.79 \cdot 10^{19} > nw_{s_1|s_0} \approx 2.21 \cdot 10^{17}$.

Now to analyse the traded energy through the pairwise interactions among agents, we plot the periodical distributions of the traded energy (settled by the \emph{contract}) over the entire simulation period of 7 days projected over a day.
Figure~\ref{fig_q_dist} illustrates the distribution through box-whisker plots. 
The flexibility of agents in exchanging energy is quite apparent during the sunny period, while in the non-sunny periods, the agents are more interested in exchanging smaller volume of energy with peers.

Figure~\ref{fig_total_trade_time} and~\ref{fig_total_trade_comp} provide insights on the impact of the trading resulting from the P2P negotiation strategy on the underlying physical network -- within the REC and through Point of Common Coupling (PCC), respectively. 
As the negotiation process requires energy to be exchanged back and forth between two prosumers, it is important to analyse the effect of that double exchange on the distribution network. Evidently, the proposed negotiation based strategy does not impose excessive strain in the network as shown in Figure~\ref{fig_total_trade_time}. 
The figure illustrates that the periodical distribution of the peak exchanged power is at its peak around the range of $\rm2.4kW$ during the sunny hours where relatively higher 
congestions are expected. Having said that, the peak is tolerable considering the size of the REC and under a safe limit of $\rm5.0kW$\footnote{considering the Dutch power system of $\rm3-phase$ $\rm16Amp$ with $230V$.}. Figure~\ref{fig_total_trade_comp} describes the relative distributions of peak power exchanged resulting from the proposed strategy and the \emph{Individual control} strategy. 
The strategy encourages exchange within REC and hence reduces the interactions with the external grid. While the average of peak power flow through PCC using the proposed strategy is $\rm0.36kW$, the same for the \emph{individual and control} strategy is $0.39kW$. The peak power, i.e. the maximum of the periodical peak flow, however, increases from $\rm2.40kW$ to $\rm6.34kW$ when the proposed strategy is utilized. Nevertheless, the power is still under the trip-limit of a PCC hosting the REC. Therefore, the strain on the network due to energy exchange is not excessive. Note that in our setting, the agents had no particular concern to minimize the peak; however, in cases where peak congestion at POC is expected agents could model a shared peak price component or the risk of peak violation in their preferences, thus explicitly considering network constraints.

\subsection{Case 3: Cooperative of 100 Agents}
In this case, we investigate the scalability of the proposed negotiation based strategy by extending Case 2's 9 agents to 100 agents. The load and generation profiles of the additional agents are copied with small variances. Figure~\ref{fig_improvement_sw_100} illustrates the relative improvements in social welfare criteria using the proposed \emph{Negotiation and control} strategy (similar to Figure~\ref{fig_improvement_sw}) when the size of the cooperative is multiplied.
As shown in the Figure, the proposed strategy improves the social welfare indices from both \emph{utilitarian} and \emph{fairness} perspectives for all agents in the cooperative. The small clusters of the scattered points are due to the small variances in the extended prosumers' profiles.

\begin{figure}[h]
\centering
\includegraphics[scale=0.35]{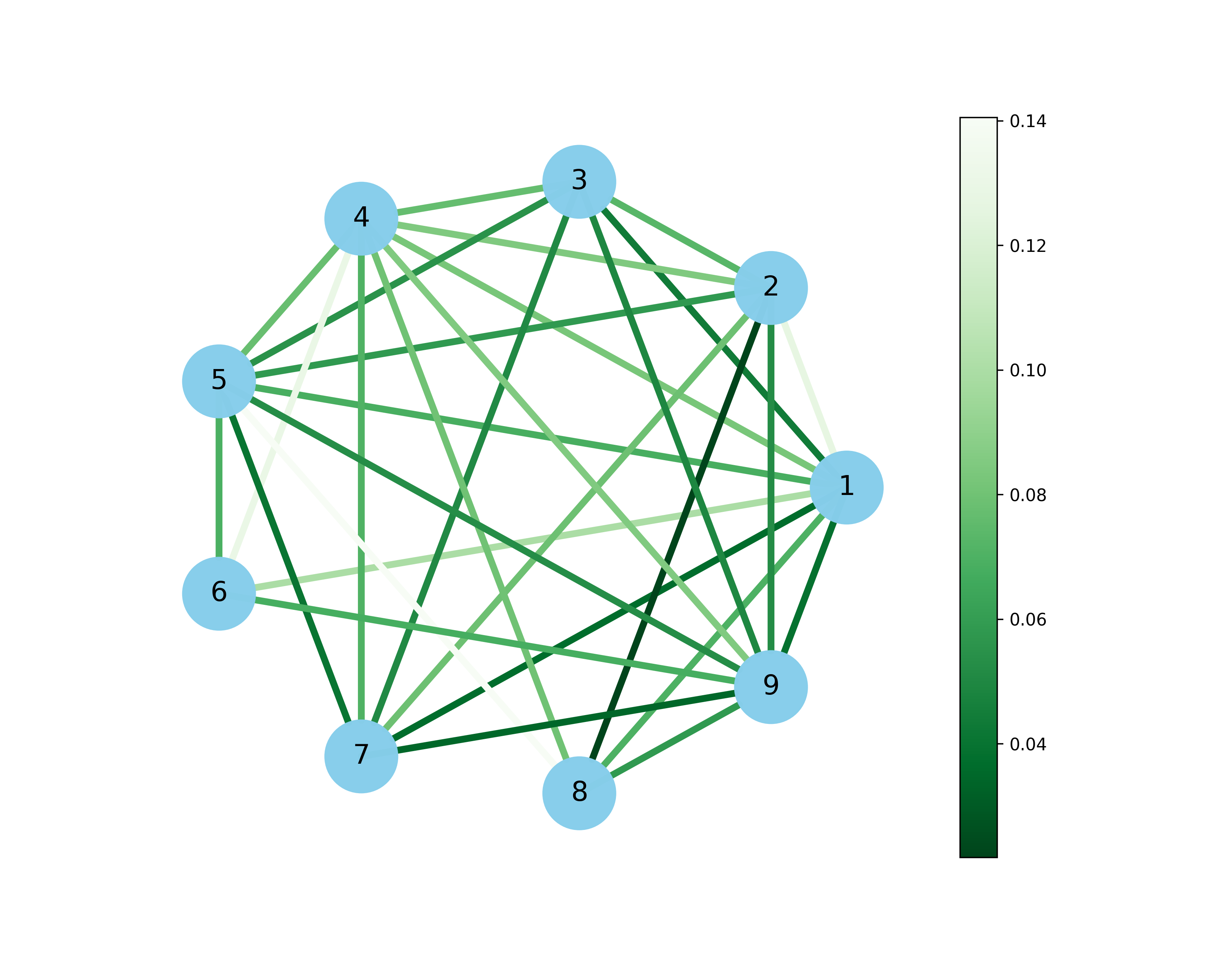}
\caption{Negotiation preference graph of the cooperative; the prosumers are plotted as the nodes. The gradient of the edges represent the strength of the P2P negotiation outcome calculated as the Euclidean distance between the \emph{agreement} and the Nash solution averaged over 4 days.}
\label{fig_pref_graph}
\end{figure}



\subsubsection{P2P interaction analyses}
In this section, we analyse the interactions between agents throughout the negotiation process. Particularly, we focus on agents' behaviours in settling contracts with their peers. 
An experiment is conducted by executing negotiation with all possible pairs throughout a particular period - e.g. 4 days. The idea is to seek for any emerging 
segmentations (or, lack thereof) among agents while identifying an appropriate set of agents who could be beneficial by pairing-up with each other to produce a successful and efficient exchange of energy.
We utilised the matrix of \emph{fair outcome} through \emph{Nash solution} to identify the strength of a pair. The \emph{fair outcome} measure is also utilised to form the basis of peer selection process (Section~\ref{sec_peer_selection}). 
As mentioned previously, the \emph{fairness} of an \emph{agreement} measured by the distance between the \emph{agreement} and the \emph{Nash solution}; the closer the \emph{agreement} is from the \emph{Nash solution}, the fairer the contract is.
Figure~\ref{fig_pref_graph} is created considering the Euclidean distance between the \emph{agreements} and the \emph{Nash solutions} averaged over a period of negotiation session ($T=\textrm{384}$, equivalent to 4 days). 
The figure highlights the quality of the pairwise interactions in achieving a fair outcome. The graph does not represent the physical network of the REC, rather it is a qualitative representation of the P2P interaction analyses of the agents. Analysing the graph, we could identify the set of prosumers that should (or not) negotiate with each other to create an optimized cooperative. For instance, agents 6, 7, and 8 should not engage in any negotiation as they never settle for any contract. The graph further strengthens the importance of having an intelligent peer selection procedure as opposed to a randomized strategy for the same.

\section{Conclusion}
\label{sec_conclusion}
A residential cooperative potentially exhibits inefficiencies due to renewable power integration and uncoordinated activations of locally owned distributed energy resources of heterogeneous prosumers.
Automated negotiation -- a natural model of interaction -- has the ability to alleviate these inefficiencies by accommodating the heterogeneous preferences of prosumers in joint decision making. In this paper, we present a peer-to-peer automated \emph{bilateral negotiation} strategy for \emph{energy contract} settlement between prosumers. The prosumers jointly seek for an \emph{agreement} on \emph{energy contracts/loans} -- consisting of energy volume to be exchanged and the return time of the exchanged energy -- that maximises their preferences by evaluating the realized energy profiles and the consequent flexibility dispatch. 
Although we consider a predefined set of criteria for the agents to have the preferences on, in reality, the agents may have a diverse set of mutually exclusive constraints that shape up their personal preferences. The proposed negotiation strategy allows the agents to effortlessly stack-up those local constraints weighing by preferences while settling for the \emph{contracts}. Additionally, the strategy inherently follows a \emph{closed-loop} solution framework and thereby alleviating the uncertainties imposed by local load and generation profiles.
The prosumers utilise an intelligent peer selection strategy that increases the quality of an \emph{agreement} and the likelihood of the negotiation being successful. The proposed negotiation based strategy has the potential to identify the group of negotiation-compatible prosumers.
The strategy is applied to real energy profiles, and results in an improved \emph{utilitarian} social welfare as well as improved fairness w.r.t. \emph{Nash} social welfare; which is remarkable considering that the allocations are achieved from single pairwise interactions amongst prosumers.

In this paper, we assume the weights an agent places on the criteria to be predefined, whereas in practice, an agent may be uncertain about the preferences and may need to elicit them from prosumers in a cost-effective way~\citep{BaarslagVisionary2017,Long2018}. Future work may investigate the case where the agents exhibit uncertainty over the preferences and are required to negotiate successfully with partial preferences. Additionally, we do not explicitly consider the physical nature of the distribution system and hence refrain to perform a power-flow analysis on the underlying distribution network. Such a consideration is really critical in analysing the applicability of the proposed P2P negotiation and consequent exchange in real microgrid setting - particularly, the real-time effect of power-flow, both active and reactive, resulting from the \emph{exchange and return} nature of the proposed method. Moreover, the impact of such exchange on the voltage and frequency regulation should need to be carefully analysed and calibrated.
However, we statistically showed that the resulting exchange due to the allocation determined by the proposed method do not stress out the network, and hence the proposed method may be implemented in the real system. Having said that, in a follow-up research, we will investigate the effect of the proposed exchange considering the physicality of the distribution network.

\section*{Acknowledgment}
The Fraunhofer Institute for Industrial Mathematics ITWM, Kaiserslautern, has kindly provided power time series data for residential load and PV generation, which is underlying the numerical evaluation of the proposed methodology. 
This research has received funding through the ERA-Net Smart Grid Plus project Grid-Friends (with support from the European Union’s Horizon 2020
research and innovation programme) and the Veni research programme with project number 639.021.751, which is financed by the Netherlands Organisation for Scientific Research (NWO). Part of this research is conducted at the Energy Transition Hub research centre at the University of Melbourne, Australia.

\bibliographystyle{elsarticle-num}

\end{document}